\documentclass[10pt, letterpaper]{article}
\usepackage[pdftex]{graphicx}
\DeclareGraphicsExtensions{.pdf,.jpeg,.png, .jpg}

\usepackage{amsmath}
\usepackage{amssymb}
\usepackage{algorithm}
\usepackage{algorithmic}
\usepackage{graphicx}
\usepackage{color}
\usepackage{array,booktabs,calc}
\usepackage{amsmath} 
\usepackage{amssymb} 
\usepackage{mathrsfs}
\usepackage{theorem}
\PassOptionsToPackage{hyphens}{url}\usepackage{hyperref}

\newcommand{\be}[1]{\begin{equation}\label{#1}}
\newcommand{\benon}{\begin{equation*}}  
\newcommand{\bemuln}[1]{\begin{multline}\label{#1}}
\newcommand{\bemul}{\begin{multline*}}
\newcommand{\bee}{\begin{eqnarray*}}
\newcommand{\eee}{\end{eqnarray*}}
\newcommand{\been}[1]{\begin{eqnarray}\label{#1}}
\newcommand{\eeen}{\end{eqnarray}}
\newcommand{\began}[1]{\begin{gather}\label{#1}}
\newcommand{\bega}{\begin{gather*}}
\newcommand{\bealn}[1]{\begin{align}\label{#1}}
\newcommand{\beal}{\begin{align*}}
\newcommand{\bealatn}[2]{\begin{alignat}{#1}\label{#2}}
\newcommand{\bealat}{\begin{alignat*}}
\newcommand{\bexalatn}[1]{\begin{xalignat}\label{#1}}
\newcommand{\bexalat}{\begin{xalignat*}}

\newcommand{\qed}{\newline \mbox{ } \hfill 
            \rule[-1pt]{2.5mm}{2.5mm}\par\vskip 10pt }

\newcommand{\pf}{\vskip 5 pt \noindent {\bf{Proof: }}}

\newcommand{\mb}{\mathbf}

{\theoremstyle{plain} \newtheorem{thm}{Theorem}[section]

\newtheorem{prop}[thm]{Proposition}

\newtheorem{cor}[thm]{Corollary}
}
{\theoremstyle{break} \theorembodyfont{\it}

 }

\def\bx{{\mathbf x}}  

\def\bz{{\mathbf z}}

\def\bM{{\mathbf M}}

\def\texitem#1{\par\smallskip\noindent\hangindent 25pt
               \hbox to 25pt {\hss #1 ~}\ignorespaces}

\newcommand{\bzero}{{\mathbf{0}}}

\newcommand{\scrP}{\mathcal{P}}

\newcommand{\scrS}{\mathcal{S}}

\newcommand{\bbeta}{\boldsymbol{\beta}}

\newcommand{\btheta}{\boldsymbol{\theta}}

\begin{document}
\title{Predicting Chronic Disease Hospitalizations from Electronic
  Health Records: An Interpretable Classification Approach \thanks{To
    appear, {\em Proceedings of the IEEE}. 
    Research partially supported by the NSF under grants CNS-1645681,
    CCF-1527292, IIS-1237022, and IIS-1724990, by the ARO under grant
    W911NF-12-1-0390, by the NIH under grant 1UL1TR001430 to the
    Clinical \& Translational Science Institute at Boston University,
    and by the Boston University Digital Health Initiative.}}

\author{Theodora S. Brisimi, Tingting Xu, Taiyao Wang, Wuyang Dai,
  \thanks{$\dagger$ Center for Information and Systems Eng., Boston University,
  e-mail: 
  {\tt \{tbrisimi, tingxu, wty, wydai\}@bu.edu}.}
  \\
William G. Adams, 
\thanks{$\ddagger$ Boston Medical Center, 850 Harrison Avenue
5th Floor, Boston, MA 02118 e-mail:
  {\tt Bill.Adams@bmc.org}.}
and Ioannis Ch. Paschalidis  
\thanks{$\S$ Department of
  Electrical and Computer Eng., Division of Systems Eng., and Dept. of
  Biomedical Eng., Boston
  University, 8 St. Mary's St., Boston, MA 02215, e-mail: {\tt
    yannisp@bu.edu}, \url{http://sites.bu.edu/paschalidis/}.}}

\maketitle

\begin{abstract}
  Urban living in modern large cities has significant adverse effects on
  health, increasing the risk of several chronic diseases. We focus on
  the two leading clusters of chronic disease, heart disease and
  diabetes, and develop data-driven methods to predict hospitalizations
  due to these conditions. We base these predictions on the patients'
  medical history, recent and more distant, as described in their
  Electronic Health Records (EHR). We formulate the prediction problem
  as a binary classification problem and consider a variety of machine
  learning methods, including kernelized and sparse Support Vector
  Machines (SVM), sparse logistic regression, and random forests. To
  strike a balance between accuracy and interpretability of the
  prediction, which is important in a medical setting, we propose two
  novel methods: $K$-LRT, a likelihood ratio test-based method, and a
  Joint Clustering and Classification (JCC) method which identifies
  hidden patient clusters and adapts classifiers to each cluster. We
  develop theoretical out-of-sample guarantees for the latter method. We
  validate our algorithms on large datasets from the Boston Medical
  Center, the largest safety-net hospital system in New England.
\end{abstract}


\section{Introduction} \label{sec:intro}

Living in modern large cities is impacting our health in many different
ways~\cite{cityhealth15}. Primarily due to: $(i)$ stress associated with
fast-paced urban life, $(ii)$ a sedentary lifestyle due to work
conditions and lack of time, $(iii)$ air pollution, and $(iv)$ a
disproportionate number of people living in poverty, urban populations
face an increased risk for the development of chronic health
conditions~\cite{yach2004global}. For example, according to the World
Health Organization~\cite{who-facts-pollution}, ambient (outdoor air)
pollution was estimated in 2012 to cause 3 million premature deaths
worldwide per year; this mortality is due to exposure to small
particulate matter of 10 microns or less in diameter (PM10), which cause
cardiovascular, respiratory disease, and cancers. In fact, the vast
majority (about 72\%) of these air pollution-related premature deaths
were due to ischemic heart disease and strokes.

There is an increasing percentage of the world population facing the adverse
health effects of urban living. Specifically, according to the United
Nations~\cite{un-desa-14}, 54\% of the earth's population resides in urban
areas, a percentage which is expected to reach 66\% by 2050. It becomes
evident that the health of citizens should become an important priority in the
emerging smart city agenda~\cite{smartcity-health-16}. To that end, smart
health care --``smart health'' as it has been called-- involves the use
of ehealth and mhealth systems, intelligent and connected medical devices, and
the implementation of policies that encourage health, wellness, and
well-being~\cite{smartcitfrosul}. It is estimated that by 2020 the smart city
market will be worth about \$1.5 trillion, with smart health
corresponding to 15\% of that amount~\cite{smartcitfrosul}. Additional
potential actions smart cities can adopt include ways to improve city life,
reduce congestion and air pollution levels, discourage the use of tobacco
products and foods high in fat and sugar which increase the risk of chronic
diseases, and improve access to health care. Without overlooking the
importance of all these population-level measures, our work aims at enabling
{\em personalized} interventions using an algorithmic data-driven approach.

Through smart health, smart cities and governments aim at improving the
quality of life of their citizens.  In the state of Massachusetts, the
MassHealth program --a combination of Medicaid and the Children's Health
Insurance Program-- provides health insurance for 1.9 million
Massachusetts residents, children in low-income households, low-wage
workers, elders in nursing homes, people with disabilities, and others
with very low incomes who cannot afford insurance~\cite{MassHealthcover,
  MassHealthCost}. The state's fiscal year 2018 budget includes
approximately $\$16.6$ billion for MassHealth, which is around $37\%$ of
the total state budget~\cite{MassHealthCost}. Clearly, this is a
substantial share of the budget. Consequently, if health care costs can
be lowered through smart health, more resources will become available
for many other services smart cities can offer. Conversely, if other
aspects of smart cities can be improved, the adverse health effects of
urban living can be reduced, thus lowering health care costs. This
suggests a beneficial feedback loop involving smart health and
non-health-related smart city research.

Health care is also, unquestionably, an important national and global
economic issue. In 2013, the United States (U.S.) spent about \$3
trillion on health care, which exceeded 17\% of its
GDP~\cite{kayyali2013big}.  The World Health Organization estimates that
healthcare costs will grow to 20\% of the U.S.\ GDP (nearly \$5
trillion) by 2021 \cite{center2012driving}, especially with civilization
diseases (or else called lifestyle diseases), like diabetes, coronary
heart disease and obesity, growing.

Our goal in this paper is to explore and develop {\em predictive
  analytics} aiming at predicting hospitalizations due to the two
leading chronic diseases: heart disease and diabetes. Prediction,
naturally, is an important first step towards prevention. It allows
health systems to target individuals most in need and to use (limited)
health resources more effectively. We refer to \cite{pas-hbr-17} for a
general discussion of the benefits, and some risks, associated with the
use of health analytics. We seek to predict hospitalizations based on
the patients' {\em Electronic Health Records (EHR)} within a year from
the time we examine the EHR, so as to allow enough lead time for
prevention. What is also critical is that our methods provide an
interpretation (or explanation) of the predictions. Interpretability
will boost the confidence of patients and physicians in the results,
hence, the chance they will act based on the predictions, and provide
insight into potential preventive measures. It is interesting that
interpretability is being increasingly recognized as important; for
instance recent European Union legislation~\cite{eu-gdpr-16} will
enforce a citizen's right to receive an explanation for algorithmic
decisions.

Our focus on heart disease and diabetes is deliberate. Diseases of the
heart have been consistently among the top causes of death. In the U.S.,
heart disease is yearly the cause of one in every four deaths, which
translates to 610,000 people~\cite{cdcp2015heart}. At the same time,
diabetes is recognized as the world's fastest growing chronic
condition~\cite{global-diabetes}. One in eleven adults has diabetes
worldwide (415 million) and 12\% of global health expenditures is spent
on diabetes (\$673 billion)~\cite{idf2015diabetes}. In the U.S. alone,
29.1 million people or 9.3\% of the population had diabetes in
2012~\cite{cdcp2014diabetes}.

Our interest in hospitalizations is motivated by
\cite{jiang2009nationwide}, which found that nearly \$30.8 billion in
hospital care cost during 2006 was preventable. Heart diseases and
diabetes were the leading contributors accounting, correspondingly, for
more than \$9 billion, or about 31\%, and for almost \$6 billion, or
about 20\%. Clearly, even modest percentage reductions in these amounts
matter.

An important enabler of our work is the increasing availability of
patients' EHRs. The digitization of patients' records started more than
two decades ago. Widespread adoption of EHRs has generated massive
datasets. 87\% of 
U.S. office-based physicians were using EHRs by the end of 2015, up from
42\% in 2008~\cite{EHRstats2016}. EHRs have found diverse uses
\cite{ludwick2009adopting}, e.g., in assisting hospital quality
management~\cite{takeda2003health}, in detecting adverse drug
reactions \cite{hannan1999detecting}, and in general primary care
\cite{wang2003cost}.

\subsection{Contributions and Organization}

Our algorithmic approach towards predicting chronic disease
hospitalizations employs a variety of methods, both already
well-established, as well as novel methods we introduce, tailored to
solve the specific medical problem. We formulate the problem as a binary
classification problem and seek to differentiate between patients that
will be hospitalized in a target year and those who will not. We review
related work in Section~\ref{sec:related}.
Section~\ref{sec:basemethods} explores baseline methods that separate
the two classes of samples (patients) using a single classifier. We
evaluate their performance in terms of prediction accuracy and
interpretability of the model and the results. Baseline methods include
linear and kernelized Support Vector Machines (SVM), random forests, and
logistic regression.  We also develop a novel likelihood ratio-based
method, $K$-LRT, that identifies the $K$ most significant features for
each patient that lead to hospitalization. Surprisingly, this method,
under a small value of $K$, performs not substantially worse than more
sophisticated classifiers using all available features. This suggests
that in our setting, a sparse classifier employing a handful of features
can be very effective. What is more challenging is that the
``discriminative'' features are not necessarily the same for each
patient.

Motivated by the success of sparse classifiers, in Section~\ref{sec:acc}
we seek to {\em jointly} identify clusters of patients who share the
same set of discriminative features and, at the same time, develop
per-cluster sparse classifiers using these features. Training such
classifiers amounts to solving a non-convex optimization problem. We
formulate it as an integer programming problem; which limits its use to
rather smaller instances (training sets). To handle much larger
instances we develop a local optimization approach based on alternating
optimization. We establish the convergence of this local method and
bound its Vapnik-Chervonenkis (VC) dimension; the latter bound leads to
out-of-sample generalization guarantees.

In Section \ref{sec:data}, we provide a detailed description of the two
datasets we use to evaluate the performance of the various algorithms.
One dataset concerns patients with heart-related diseases and the other,
patients with diabetes. The data have been made available to us from the
Boston Medical Center (BMC) -- the largest safety-net hospital in New
England. We define the performance metrics we use in
Section~\ref{sec:per}. We report and discuss our experimental settings
and results in Section \ref{sec:results} and we present our conclusions
in Section \ref{sec:concl}.

\textbf{Notation:} All vectors are column vectors. For economy of space, we
write $\bx = \left(x_1, \ldots, x_{\text{dim}(\bx)}\right)$ to denote the
column vector $\bx$, where $\text{dim}(\bx)$ is the dimension of $\bx$. We use
$\bzero$ and $\mb{1}$ for the vectors with all entries equal to zero and one,
respectively.  We denote by $\mathbb{R}_+$ the set of all nonnegative real
numbers. $\bM \geq \textbf{0}$ (resp., $\bx \geq \textbf{0}$) indicates that
all entries of a matrix $\bM$ (resp., vector $\bx$) are nonnegative. We use
``prime'' to denote the transpose of a matrix or vector and $\left|
\mathcal{D} \right|$ the cardinality of a set $\mathcal{D}$.  Unless otherwise
specified, $\|\cdot\|$ denotes the $\ell_2$ norm and  $\|\cdot\|_1$ the
$\ell_1$ norm.

\section{Related Work} \label{sec:related}

To the best of our knowledge, the problem of chronic disease
hospitalization prediction using machine learning methods is novel. A
closely related problem, which has received a lot of attention in the
literature, is the re-hospitalization prediction, since around 20\% of
all hospital admissions occur within 30 days of a previous
discharge. Medicare penalizes hospitals that have high rates of
readmissions for some specific conditions that now include patients with
heart failure, heart attack, and pneumonia. Examples of work on this
problem include \cite{hosseinzadeh2013assessing},
\cite{zolfaghar2013big}, \cite{strack2014impact} and
\cite{caruana2015intelligible}.

Other related problems considered in the literature are: predicting the
onset of diabetes using artificial neural networks
\cite{pradhan2011predict}; developing an intelligent system that
predicts, using data-mining techniques, which patients are likely to be
diagnosed with heart disease \cite{palaniappan2008intelligent}; and
using data-mining techniques to predict length of stay for cardiac
patients (employing decision trees, SVM, and artificial neural networks)
\cite{hachesu2013use}, or for acute pancreatitis (using artificial neural
networks) \cite{pofahl1998use}.

We should also mention the Heritage Health Prize, a competition by
Kaggle, whose goal was to predict the length of stay for patients who
will be admitted to a hospital within the next year, using insurance
claims data and data-mining techniques \cite{heritage2011heritage}.

\section{Baseline Methods and $K$-LRT} 
\label{sec:basemethods}  

In this section we outline several baseline classification methods we
use to predict whether patients will be hospitalized in a target year,
given their medical history. 

In medical applications, accuracy is important, but also
interpretability of the predictions is
indispensable~\cite{vellido2012making}, strengthening the confidence of
medical professionals in the results. Sparse classifiers are
interpretable, since they provide succinct information on few dominant
features leading to the prediction \cite{lee2006efficient}. Moreover,
medical datasets are often imbalanced since there are much fewer
patients with a condition (e.g., hospitalized) vs.\ ``healthy''
individuals (non-hospitalized). This makes it harder for supervised
learning methods to learn since a training set may be dominated by
negative class samples. Sparsity, therefore, is useful in this context
because there are fewer parameters in the classifier one needs to learn.
In this light, we experiment with sparse versions of various
classification methods and show their advantages. While harder to
interpret than linear and sparse algorithms, ensemble methods that build
collections of classifiers, such as random forests, can model nonlinear
relationships and have been proven to provide very accurate models for
common health care problems~\cite{casanova2014application}, including
the one we study in this paper.

The last method we present in this section is an adaptation of a
likelihood ratio test, designed to induce sparsity of the features used
to make a prediction. All but the last method fall into the category of
discriminative learning algorithms, while the last one is a generative
algorithm. Discriminative algorithms directly partition the input space
into label regions without modeling how the data are generated, while
generative algorithms assume a model that generates the data, estimate
the model's parameters and use it to make classification decisions. Our
experiment results show that discriminative methods are likely to give
higher accuracy, but generative methods provide more interpretable
models and results~\cite{vapnik,ng2002discriminative}. This is the
reason we experiment with methods from both families and the trade-off
between accuracy and interpretability is observed in our results.

\subsection{RBF, Linear \& Sparse Linear Support Vector Machines} 

A Support Vector Machine (SVM) is an efficient binary
classifier~\cite{cortes1995support}. The SVM training algorithm seeks a
separating hyperplane in the feature space, so that data points from the two
different classes reside on different sides of that hyperplane. We can
calculate the distance of each input data point from the hyperplane. The
minimum over all these distances is called {\em margin}. The goal of SVM is to
find the hyperplane that has the maximum margin. In many cases, however, data
points are neither linearly nor perfectly separable. So called {\em
  soft-margin} SVM, tolerates misclassification errors and can leverage kernel
functions to ``elevate'' the features into a higher dimensional space where
linear separability is possible ({\em kernelized
  SVMs})~\cite{cortes1995support}.

Given our interest in interpretable, hence sparse, classifiers we
formulate a Sparse version of Linear SVM (SLSVM) as follows. We are
given training data $\bx_i\in\mathbb{R}^D$ and labels $y_i\in\{-1,1\}$,
$i=1,\dots, n$, where $\bx_i$ is the vector of features for the $i$th
patient and $y_i=1$ (resp., $y_i=-1$) indicates that the patient will
(resp., not) be hospitalized. We seek to find the classifier $(\bbeta,
\beta_0)$, $\bbeta \in\mathbb{R}^D, \beta_0\in\mathbb{R}$, by solving:
\begin{equation}
\begin{array}{ll}
\min\limits_{\bbeta,\beta_0,\xi_i} & \frac{1}{2} \|\bbeta\|^2+C
\sum_{i=1}^{n}\xi_i+\rho\|\bbeta\|_1\\ 
\text{s.t.} & \xi_i \geq 0, \quad \forall i, \\
& y_i(\bx_i'\bbeta+\beta_0)\geq1-\xi_i, \quad \forall i,\\
\end{array}
\label{svm_sparse}
\end{equation}
where $\xi_i$ is a misclassification penalty. The first term in the
objective has the effect of maximizing the margin. The second objective
term minimizes the total misclassification penalty. The last term,
$\|\bbeta\|_1$, in the objective, imposes sparsity in the feature vector
$\bbeta$, thus allowing only a sparse subset of features to contribute
to the classification decision. The parameters $C$ and $\rho$ are
tunable parameters that control the relative importance of the
misclassification and the sparsity terms, respectively, compared to each
other and, also, the margin term. When $\rho=0$, the above formulation
yields a standard linear SVM classifier.

A linear SVM finds a linear hyperplane in the feature space and can not
handle well cases where a nonlinear separating surface between classes
is more appropriate. To that end, kernel functions are being used that
map the features to a higher dimensional space where a linear hyperplane
would be applicable. In the absence of the sparse-inducing $\ell_1$-norm
term, kernelized SVMs use $K(\bx_i,\bx_j) = \phi(\bx_i)' \phi(\bx_i)$ as
a kernel for some feature mapping function $\phi$ and solve an
optimization problem that is based on the dual problem to
(\ref{svm_sparse}) to find an optimal $(\bbeta, \beta_0)$.  In our
application, we will employ the widely used Radial Basis Function (RBF)
$K(\bx_i,\bx_j)= \exp (-\|\bx_i-\bx_j \|^2/2\sigma^2)$
\cite{scholkopf1997comparing} as the kernel function in our experiments.

\subsection{Random Forests}

Bagging (or bootstrap aggregating) is a technique for reducing the
variance of an estimated predictor by averaging many noisy but
approximately unbiased models. A random forest is an ensemble of
de-correlated trees \cite{friedman2001elements}. Each decision tree is
formed using a training set obtained by sampling (with replacement) a
random subset of the original data. While growing each decision tree,
random forests use a random subset of the set of features (variables) at
each node split. Essentially, the algorithm uses bagging for both trees
and features.  Each tree is fully grown until a minimum size is reached,
i.e., there is no pruning. While the predictions of a single tree are
highly sensitive to noise in its training set, the average of many trees
is not, as long as the trees are not correlated. Bagging achieves
de-correlating the trees by constructing them using different training
sets. To make a prediction at a new sample, random forests take the
majority vote among the outputs of the grown trees in the
ensemble. Random forests run very efficiently for large datasets, do not
have the risk of overfitting (as, e.g.,
AdaBoost~\cite{freund1995desicion}, a boosting method) and can handle
datasets with imbalanced classes. The number of trees in the ensemble is
selected through cross-validation.

\subsection{Sparse Logistic Regression} \label{sec:LR}

Logistic Regression (LR) \cite{bishop2006pattern} is a linear classifier
widely used in many classification problems. It models the posterior
probability that a patient will be hospitalized as a logistic function
of a linear combination of the input features, with parameters $\btheta$
that weigh the input features and an offset $\theta_0$. The parameters
of the model are selected by maximizing the log-likelihood using a
gradient method. For the test samples, decisions are made by
thresholding the log-likelihood ratio of the positive (hospitalized)
class over the negative class. Logistic regression is popular in the
medical literature because it predicts a probability of a sample
belonging to the positive class. Here, we use an $\ell_1$-regularized
(sparse) logistic regression \cite{lee2006efficient,
  pudil1994floating,street-bump-16}, which adds an extra penalty term
proportional to $\|\btheta\|_1$ in the log-likelihood. The motivation is
to induce sparsity, effectively ``selecting'' a sparse subset of
features.  More specifically, we solve the following convex problem
using a gradient-type method:
\begin{equation}
\begin{array}{ll}
\min\limits_{\btheta,\theta_0} & \sum_{i=1}^{n} (-\log p(y_i|\bx_i;
\btheta,\theta_0))  +
\lambda \|\btheta\|_1\\ 
\end{array}
\label{log_sparse}
\end{equation}
where the likelihood function is given by
\begin{align*}
p(y_i=1|\bx_i; \btheta,\theta_0) = & \frac{1}{1+e^{-\theta_0-\btheta'\bx_i}}\\
  = & 1-p(y_i=-1|\bx_i; \btheta,\theta_0),
\end{align*}
and $\lambda$ is a tunable parameter controlling the sparsity
term. Setting $\lambda=0$, we obtain a standard logistic regression
model.

\subsection{$K$-Likelihood Ratio Test}

The {\em Likelihood Ratio Test (LRT)} is a naive Bayes classifier and assumes
that individual features (elements) of the feature vector
$\bx=(x_1,\ldots,x_D)$ are independent random
variables~\cite{dai-bri-pas-prediction-ijmi-14}. The LRT algorithm
empirically estimates the distribution $p(x_j|y)$ of each feature $j$ for the
hospitalized and the non-hospitalized class. Given a new test sample
$\bz=(z_1, z_2, \cdots, z_D)$, LRT calculates the two likelihoods $p(\bz|y=1)$
and $p(\bz|y=-1)$ and then classifies the sample by comparing the ratio
\[
\frac{p(\bz|y=1)}{p(\bz|y=-1)}= \prod_{j=1}^{D} \frac{p(z_j|y=1)}{p(z_j|y=-1)}
\]
to a threshold. In our variation of the method, which we will call
$K$-LRT,~\footnote{$K$-LRT was first proposed in
  ~\cite{dai-bri-pas-prediction-ijmi-14} and was applied only to a
  heart-disease dataset.} instead of taking into account the ratios of
the likelihoods of all features, we consider only the $K$ features with
the largest ratios. We consider only the largest ratios because they
correspond to features with a strong hospitalization ``signal.''  On the
other hand, we do not consider features with the smallest ratios because
they could be due to the imbalance of the dataset which has much more
non-hospitalized than hospitalized patients.

The optimal $K$ can be selected using cross-validation from a set of
pre-defined values, that is, as the value with the best classification
performance in a validation set. The purpose of this ``feature
selection'' is again sparsity, that is, to identify the $K$ most
significant features for each individual patient. Thus, each patient is
actually treated differently and this algorithm provides
interpretability as to why a specific classification decision has been
made for each individual patient.

\section{Joint Clustering and Classification (JCC)} \label{sec:acc}

In this section, we introduce a novel {\em Joint Clustering and
  Classification} method. The motivation comes from the success of
$K$-LRT, which we will see in Section~\ref{sec:results}. Since $K$-LRT
selects a sparse set of features for each patient, it stands to reason
that there would be clusters of patients who share the same
features. Moreover, since $K$-LRT uses the $K$ largest likelihood
ratios, feature selection is more informative for patients that are
hospitalized (positive class). This is intuitive: patients are
hospitalized for few underlying reasons while non-hospitalized patients
appear ``normal'' in all features associated with a potential future
hospitalization.

\begin{figure}[ht]
	\begin{centering}
		\includegraphics[width=0.5\columnwidth]{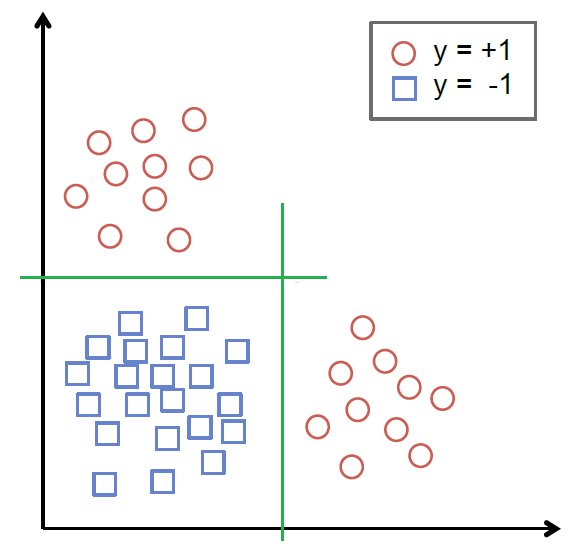}
		\caption{The positive class contains two clusters and each
                  cluster is linearly separable from the negative class.}
		\label{figure:asymmetricDemo}
	\end{centering}
\end{figure}

To reflect this reasoning, we consider a classification problem in which
the positive class consists of multiple clusters, whereas negative class
samples form a single cluster. It is possible to extend our framework
and consider a setting where clustering is applied to both the positive
and the negative class. However, because our results are satisfactory
and to avoid further increasing complexity, we do not pursue this
direction in this work. We assume that for each (positive class) cluster
there is a sparse set of discriminative dimensions, based on which the
cluster samples are separated from the negative
class. Fig.~\ref{figure:asymmetricDemo} provides an illustration of this
structure. The different clusters of patients are naturally created
based on age, sex, race or different diseases. From a learning
perspective, if the hidden positive groups are not predefined and we
would like to learn an optimal group partition in the process of
training classifiers, the problem could be viewed as a combination of
clustering and classification.  Furthermore, with the identified hidden
clusters, the classification model becomes more interpretable in
addition to generating accurate classification labels. A preliminary
theoretical framework for JCC appeared in our conference paper
\cite{xu2016joint}, but without containing all detailed proofs of the
key technical results and with very limited numerical evaluation.

\subsection{An integer programming formulation} 

We next consider a joint cluster detection and classification problem
under a Sparse Linear SVM (SLSVM) framework.  Let $\mathbf{x}_{i}^{+}$
and $\mathbf{x}_{j}^{-}$ be the $D$-dimensional positive and negative
class data points (each representing a patient), and $y_{i}^{+} = 1, \,
\forall i,$ $y_{j}^{-} = -1, \, \forall j$, the corresponding labels,
where $i\in \{1,2,\ldots,N^{+}\}$ and $j\in \{1,2,\ldots,N^{-}\}$.
Assuming $L$ hidden clusters in the positive class, we seek to discover:
$(a)$ the $L$ hidden clusters (denoted by a mapping function $l(i)=l, \
l\in \{1,2,\ldots,L\}$), and $(b)$ $L$ classifiers, one for each
cluster. Let $T^l$ be a parameter controlling the sparsity of the
classifier for each cluster $l$. We formulate the {\em Joint Clustering
  and Classification (JCC)} problem as follows:
\begin{equation}\label{opt:JCC}
\begin{array}{rl}
\min\limits_{\substack{\boldsymbol{\beta}^{l},\beta^{l}_0, l(i)\\\zeta_{j}^{l},
		\xi_{i}^{l}}} 
& \sum\limits_{l=1}^{L} \left(\frac{1}{2}\| \bbeta^{l}\|^2 
+ \lambda^{+}\sum\limits_{i:l(i)=l} \xi_{i}^{l(i)} 
+ \lambda^{-}\sum\limits_{j=1}^{N^{-}} \zeta_{j}^{l}\right)\\
\text{s.t.}  & \sum\limits_{d=1}^D |\beta^{l}_{d}| \leq T^l,\; \forall l,\\ 
&\xi_i^{l(i)} \geq 1-y^{+}_i \beta^{l(i)}_0 - \sum\limits_{d=1}^D  y^{+}_i \beta^{l(i)}_{d} x^{+}_{i,d},\;  \forall i,\\
&\zeta_j^{l} \geq 1-y^{-}_j \beta^{l}_0 - \sum\limits_{d=1}^D  y^{-}_j \beta^{l}_{d} x^{-}_{j,d},\;  \forall j, l, \\ 
&\xi_i^{l(i)}, \ \zeta_{j}^{l}\geq 0,\; \forall i, j, l.
\end{array}
\end{equation}
In the above formulation, the margin between the two classes in cluster
$l$ is equal to $2/\|\bbeta^{l}\|$, hence the first term in the
objective seeks to maximize the margin. The variables $\xi_i^{l},
\zeta_{j}^{l}$ represent misclassification penalties for the positive
and negative data points, respectively. The first constraint limits the
$\ell_1$ norm of $\bbeta^{l}$ to induce a sparse SVM for each
cluster. The second (resp., third constraint) ensures that the positive
(resp., negative) data points end up on the positive (resp. negative)
side of the hyperplane; otherwise a penalty of $\xi_i^{l}$ (resp.,
$\zeta_{j}^{l}$) is imposed; these misclassification penalties are
minimized at optimality. We use different misclassification penalties
for the positive and negative data points to accommodate a potential
imbalance in the training set between available samples; typically, we
have more negative (i.e., not hospitalized) samples. Notice that the
misclassification costs of the negative samples are counted $L$ times
because they are drawn from a single distribution and, as a result, they
are not clustered but simply copied into each cluster. The parameters
$\lambda^{-}$ and $\lambda^{+}$ control the weights of costs from the
negative and the positive samples.

As stated, problem~(\ref{opt:JCC}) is not easily solvable as it
combines the cluster allocation decisions (i.e., deciding the cluster
assignment $l(i)$ for each sample $i$) with the determination of the SVM
hyperplanes. One approach to solve JCC is shown below, where we
transform the problem into a {\em mixed integer programming} problem
(MIP) by introducing binary indicator variables to represent the cluster
assignment in JCC (each positive sample can only be assigned to one
cluster):
\begin{equation}\label{eqn:MIP}
\begin{array}{rl}
 \min\limits_{\substack{\boldsymbol{\beta}^{l}, \beta^{l}_0, z_{il}\\
      \zeta_{j}^{l}, \xi_{i}^{l}}}  
  &\sum\limits _{l=1}^{L}\left(\frac{1}{2} \|\boldsymbol{\beta}^{l}\| ^2
    +\lambda^+\sum\limits_{i=1}^{N^+}\xi^l_i
    +\lambda^-\sum\limits_{j=1}^{N^-}\zeta^l_j\right) \\
\hspace{-5pt}  \text{s.t.} &\hspace{-5pt} \sum\limits_{d=1}^{D}
|\beta_d^l| \le T^l,\; 
\forall l,\\ 
\hspace{-5pt}  &\hspace{-5pt}\xi_i^l \geq
1-y_i^+\beta_0^l-\sum\limits_{d=1}^{D}y_i^+\beta_d^l 
  x^+_{i,d}-M\sum\limits_{k \neq l} z_{ik},\ \forall i, l,\\ 
\hspace{-5pt}  &\hspace{-5pt}\zeta_j^l \geq
1-y_j^-\beta_0^l-\sum\limits_{d=1}^{D}y_j^-\beta_d^l 
  x^-_{j,d},\;  \forall j, l\\
\hspace{-5pt}  &\hspace{-5pt}\sum\limits_{l=1}^{L} z_{il}=1 ,\; \forall
i; ~ z_{il} \in \{0,1\},\; 
  \forall i, l,\\  
\hspace{-5pt}  &\hspace{-5pt}\xi_i^l, \zeta_j^l \geq 0, \forall i, j, l,
\end{array}
\end{equation}
where $z_{il}=1$ when $l(i)=l$ and $0$ otherwise (binary variables
describing the cluster assignments) and $M$ is a large positive real
number. The following proposition establishes the equivalence between
formulations (\ref{eqn:MIP}) and (\ref{opt:JCC}). The proof can be found
in Appendix~\ref{sec:appA}. 
\begin{prop} \label{prop:mip} The MIP formulation (\ref{eqn:MIP}) is
  equivalent to the original JCC formulation (\ref{opt:JCC}).
\end{prop}

In order to obtain better clustering performance, we introduce a penalty
term in the objective function seeking to minimize the intra-cluster
distances between samples, that is, making samples in the same cluster
more similar to each other. This term takes the form: $\rho
\sum_{i_1=1}^{N^+}\sum_{i_2=1}^{N^+}\sigma_{i_1i_2} \|
\boldsymbol{x}_{i_1}^+ - \boldsymbol{x}_{i_2}^+\|^2$, where
\[
\sigma_{i_1i_2}=\begin{cases} 1, \ \text{if} \ x_{i_1}^+ \ \text{and} \
x_{i_2}^+ \ \text{belong to the same cluster}, \\0, \
\text{otherwise.} \end{cases} 
\]   
For $\sigma$ to comply with this definition, we need to impose the
constraint 
\[ 
z_{i_1l} + z_{i_2l} - \sigma_{i_1i_2} \leq 1,\; \forall \
i_1\neq i_2, l\ \text{and}\ \sigma_{i,j}\in\{0,1\}.
\]

The MIP approach presented above comes in a compact form, solves jointly
the clustering and the classification problem, and exhibits good
performance on small-scale problems. However, there are no general
polynomial-time algorithms for solving MIPs, thus, making it problematic
for large datasets that are most common in practice.  This motivates us
to develop the following {\em Alternating Clustering and Classification
  (ACC)} approach, which does not suffer from these limitations.

\subsection{An alternating optimization approach}

The idea behind ACC is to alternately train a classification model and
then re-cluster the positive samples, yielding an algorithm which scales
well and also, as we will see, comes with theoretical performance
guarantees.

Given cluster assignments $l(i)$ for all positive training samples $i$,
the JCC problem (\ref{opt:JCC}) can be decoupled into $L$ separate
quadratic optimization problems, essentially solving an SVM training
problem per cluster. Our alternating optimization approach, summarized
in Algorithms~\ref{alg:ACCnonsymAlgo}--\ref{alg:reclustering2},
consists of two major modules: $(i)$ training a classifier for each
cluster and $(ii)$ re-clustering positive samples given all the
estimated classifiers. 

The process starts with an initial (e.g., random or using some
clustering algorithm) cluster assignment of the positive samples and
then alternates between the two
modules. Algorithm~\ref{alg:ACCnonsymAlgo} orchestrates the alternating
optimization process; given samples' assignment to clusters, it obtains
the optimal per-cluster SLSVM classifiers and calls the re-clustering
procedure described in Algorithm~\ref{alg:reclustering2}.

Algorithm~\ref{alg:reclustering2} uses the computed $L$ classifiers and
assigns a positive sample $i$ to the cluster $l$ whose classification
hyperplane is the furthest away from the sample $i$, that is, whose classifier
better separates sample $i$ from the negative class. Notice that the
re-clustering of the positive samples is based on $\mathcal{C}$, a subset of
$\{1,\dots,D\}$, which is a set of selected features that allows us to
select which features are important in cluster discrimination so that the
  identified clusters are more interpretable. In a notational remark, we
denote $\mathbf{x}^{+}_{i, \mathcal{C}} $ (resp., $\mathbf{x}_{\mathcal{C}}$)
as the projection of the $D$-dimensional feature vector $\mathbf{x}^{+}_{i}$
(resp., $\mathbf{x}$) on the subset $\mathcal{C}$. We also impose the
constraint (\ref{eqn:impconstr}) in Algorithm~\ref{alg:reclustering2}, which
is necessary for proving the convergence of ACC.
\begin{algorithm}[htb]
	\caption{ACC Training}
	\label{alg:ACCnonsymAlgo}
	\begin{algorithmic}
		\STATE {\bfseries Initialization:} \\
		Randomly assign positive class sample $i$ to cluster
                $l(i)$, for   
		$i \in \{1, \ldots, N^{+}\}$ and $l(i) \in \{1, \ldots, L\}$.
		\REPEAT
		\STATE {\bfseries Classification Step:} \\
		Train an SLSVM classifier for each cluster of positive
                samples combined 
		with all negative samples. Each classifier is the outcome of a
		quadratic optimization  
		problem (cf. (\ref{opt:SLSVM})) and provides a hyperplane
		perpendicular to $\boldsymbol{\beta}^{l}$ and a corresponding
		optimal objective value $O^{l}$.
		\STATE {\bfseries Re-clustering Step:} \\
		Re-cluster the positive samples based on the classifiers
		$\boldsymbol{\beta}^{l}$ and update the $l(i)$'s.  \vskip 0.1in
		\UNTIL{no $l(i)$ is changed or $\sum_{l} O^{l}$ is not
			decreasing.}
	\end{algorithmic}
\end{algorithm}

\begin{algorithm}[htb]
	\caption{Re-clustering procedure given classifiers}
	\label{alg:reclustering2}
	\begin{algorithmic}
		\STATE {\bfseries Input:} positive samples $\mathbf{x}_i^{+}$, classifiers $\boldsymbol{\beta}^{l}$,
		current cluster assignment which assigns sample $i$ to cluster $l(i)$.
		\FOR{all $i\in \{1,\ldots,N^{+}\}$}  
		\STATE \FOR{all $l\in \{1,\ldots,L\}$}
		\STATE calculate the projection $a^{l}_{i} $ of positive
		sample $i$ onto the classifier for cluster $l$ using
		only elements in a feature set $\mathcal{C}$:  
		$a^{l}_{i} = {\mathbf{x}^{+}_{i,
				\mathcal{C}}}^{'}
		\boldsymbol{\beta}^{l}_{\mathcal{C}}$ ; 
		\ENDFOR
		\STATE update cluster assignment of sample $i$ from
                $l(i)$  to \\ 
		$l^{*}(i) = \arg\max\limits_{l} a^{l}_{i}$, 
		subject to 
		\begin{equation}\label{eqn:impconstr}
		{\mathbf{x}^{+}_{i}}^{'} 
		\boldsymbol{\beta}^{l^{*}(i)}+\beta^{l^{*}(i)}_{0}\ 
		\geq\  {\mathbf{x}^{+}_{i}}^{'}
		\boldsymbol{\beta}^{l(i)}+\beta^{l(i)}_{0}. 
		\end{equation}
		\ENDFOR
	\end{algorithmic}
\end{algorithm}

Finally, Algorithm~\ref{alg:ACCnonsymAlgoTest} describes how ACC
classifies new samples not used in training. Specifically, it assigns a
new sample to the cluster whose classifier is furthest away from that
sample and uses the classifier of that cluster to make the classification
decision.
\begin{algorithm}[htb]
	\caption{ACC Testing}
	\label{alg:ACCnonsymAlgoTest}
	\begin{algorithmic}
		\FOR{each test sample $\mathbf{x}$}
		\STATE  Assign it to cluster $l^{*} = \arg\max\limits_{l} {\mathbf{x}_{\mathcal{C}}}^{'} 
		\boldsymbol{\beta}^{l}_{\mathcal{C}}$.
		\STATE Classify $\mathbf{x}$ with $\boldsymbol{\beta}^{l^{*}}$.
		\ENDFOR
	\end{algorithmic}
\end{algorithm}

\subsection{ACC performance and convergence
  guarantees} \label{sec:theory}  

In this subsection, we rigorously prove ACC convergence and establish
out-of-sample (in a ``test'' set not seen during training) performance
guarantees. While theoretical, such results are important because $(i)$
they establish that ACC will converge to a set of clusters and a
classifier per cluster and $(ii)$ characterize the number of samples
needed for training, as well as $(iii)$ bound out-of-sample
classification performance in terms of the in-sample performance.

We first present a result that suggests a favorable sample complexity
for SLSVM compared to the standard linear SVM. Suppose that SLSVM for
the $l$-th cluster yields $Q^l<D$ non-zero elements of
$\boldsymbol{\beta}^l$, thus, selecting a $Q^l$-dimensional subspace of
features used for classification. The value of $Q^l$ is controlled by
the parameter $T^l$ (cf. (\ref{eqn:MIP})).

As is common in the learning
literature~\cite{cristianini2000introduction}, we draw independent and
identically distributed (i.i.d.) training samples from some underlying
probability distribution. Specifically, we draw $N^-$ negative samples
from some distribution $\scrP_0$ and $N_l^+$ positive samples for
cluster $l$ from some distribution $\scrP^l_1$, where the total number
of positive and negative samples used to derive the classifier of
cluster $l$ is $N^l = N_l^+ +N^-$. Let $R^l_{N^l}$ denote the expected
training error rate and $R^l$ the expected test error (out-of-sample)
for the classifier of cluster $l$ under these distributions. The proof
of the following result is in Appendix~\ref{sec:appB}. We note that $e$
in (\ref{sample_size}) is the base of the natural logarithm.
\begin{thm}\label{thm:samplecomplexity}
  For a specific cluster $l$, suppose that the corresponding sparse
  linear SVM classifier lies in a $Q^l$-dimensional subspace of the
  original $D$-dimensional space. Then, for any $\epsilon>0$ and
  $\delta\in(0,1)$, if the sample size $N^l$ satisfies
\begin{equation} \label{sample_size}
N^l\geq \frac{8}{\epsilon^2} \left[ \log\frac{2}{\delta}
+ (Q^l+1)\log\frac{2eN^l}{Q^l+1}+ Q^l
\log\frac{e D}{Q^l} \right], 
\end{equation}
it follows that with probability no smaller than $1-\delta$,
$R^l-R^l_{N^l}\leq\epsilon$.
\end{thm}
Theorem~\ref{thm:samplecomplexity} suggests that if the training set
contains a number of samples roughly proportional to
$(Q^l+\log(1/\delta))/\epsilon^2$, then we can guarantee with probability
at least $1-\delta$ an out-of-sample error rate $\epsilon$-close to the
training error rate. In other words, sparse SVM classification requires
samples proportional to the effective dimension of the sparse classifier
and not the (potentially much larger) dimension $D$ of the feature
space.

Next we establish that the ACC training algorithm converges. The proof
is given in Appendix~\ref{sec:appC}. As a remark on convergence, it is
worth mentioning that the values $\lambda^{+}$ and $\lambda^{-}$ should
be fixed across all clusters to guarantee convergence.
\begin{thm}\label{thm:convergence}
  The ACC training algorithm (Alg.~\ref{alg:ACCnonsymAlgo}) converges
  for any set $\mathcal{C}$.
\end{thm}

The following theorem establishes a bound on the VC-dimension of the
class of decision functions produced by ACC training. As we will see,
this bound will then lead to out-of-sample performance guarantees. To state
the result, let us denote by $\mathcal{H}$ the family of
clustering/classification functions produced by ACC training. The proof
of the following theorem is in Appendix~\ref{sec:appD}. 
\begin{thm}\label{thm:VCdim}
The VC-dimension of $\mathcal{H}$ is bounded by 
\begin{equation*}
V_{ACC}\stackrel{\triangle}{=}(L+1)L(D+1) \log\left( e \frac{(L+1)L}{2} \right).
\end{equation*}
\end{thm}
Theorem \ref{thm:VCdim} implies that the VC-dimension of ACC-based
classification grows linearly with the dimension of data samples and
polynomially (between quadratic and cubic) with the number of
clusters. Since the local (per cluster) classifiers are trained under an
$\ell_{1}$ constraint, they are typically defined in a lower dimensional
subspace. At the same time, the clustering function also lies in a lower
dimensional space $\mathcal{C}$. Thus, the ``effective'' VC-dimension
could be smaller than the bound in Theorem~\ref{thm:VCdim}.

An immediate consequence of Thm.~\ref{thm:VCdim} is the following
corollary which establishes out-of-sample generalization guarantees for
ACC-based classification and is based on a result in
\cite{bousquet2004introduction} (see also Appendix~\ref{sec:appB}). To
state the result, let $N=N^+ +N^-$ the size of the training set. Let
$R_{N}$ denote the expected training error rate and $R$ the
expected test error (out-of-sample) of the ACC-based classifier. 
\begin{cor}\label{cor:guarantees}
 For any $\rho\in(0,1)$, with probability at least $1-\rho$ it holds: 
\[
R \leq R_{N} + 
2\sqrt{2\frac{V_{ACC}\log\frac{2eN}{V_{ACC}}+ \log\frac{2}{\rho}}{N}}.
\]
\end{cor}

\section{The Data}\label{sec:data}

The data we use to evaluate the various methods we presented come from
the Boston Medical Center (BMC). BMC is the largest safety-net hospital
in New England and with 13 affiliated Community Health Centers (CHCs)
provides care for about $30$\% of Boston residents. The data integrate
information from hospital records, information from the community health
centers, and some billing records, thus forming a fairly rich and
diverse dataset.

The study is focused on patients with at least one heart-related
diagnosis or procedure record in the period 01/01/2005--12/31/2010 or a
diagnosis record of diabetes mellitus between
01/01/2007--12/31/2012. For each patient in the above set, we extract
the medical history (demographics, hospital/physician visits, problems,
medications, labs, procedures and limited clinical observations) for the
period 01/01/2001--12/31/2010 and 01/01/2001--12/31/2012,
correspondingly, which includes relevant {\em medical factors} from
which we will construct a set of patient features. Data were available
both from the hospital EHR and billing
systems. Table~\ref{table:medfactors} shows the ontologies, along with the
number of factors and some examples corresponding to each of the heart
patients. Similarly, Table~\ref{table:diabmedfactors} shows the ontologies
with some examples for the diabetic patients. In these tables, ICD9
(International Classification of Diseases, 9th revision)~\cite{icd9},
CPT (Current Procedural Terminology)~\cite{cpt}, LOINC (Logical
Observation Identifiers Names and Codes)~\cite{loinc}, and MSDRG
(Medicare Severity-Diagnosis Related Group)~\cite{msdrg} are commonly
used medical coding systems for diseases, procedures, laboratory
observations, and diagnoses, respectively.

We note that some of the diagnoses and admissions in
Table~\ref{table:medfactors} are not directly heart-related, but may be
good indicators of a heart problem. Also, as expected, many of the
diagnoses and procedures in Table~\ref{table:diabmedfactors} are direct
complications due to diabetes. Diabetes-related admissions are not
trivially identifiable, and are revealed through the procedure described
in the next subsection. Overall, our heart dataset contains 45,579
patients and our diabetes dataset consists of 33,122 patients after
preprocessing, respectively. Among these patients, 3,033 patients in the
heart dataset and 5,622 patients in the diabetes dataset are labeled as
hospitalized in a target year. For each dataset we randomly select 60\%
of the patients for training and keep the remaining 40\% of the patients
for testing.

Our objective is to leverage past medical factors for each patient to
predict whether she/he will be hospitalized or not during a {\em target}
year which, as we explain below, could be different for each patient.

\begin{table}[hbt]
\caption{Medical Factors in the Heart Diseases Dataset.}
\label{table:medfactors}
\begin{center}
\begin{tabular}{>{\centering\arraybackslash}m{1.5cm} >{\centering\arraybackslash}m{1.8cm} m{9cm}} 
	\toprule
	\textbf{Ontology} & \textbf{Number of Factors} & \textbf{Examples}\\ 
	\midrule
	Demographics & 4 & Sex, Age, Race, Zip Code  \\ 
[0.5em]
	Diagnoses & 22 & e.g., Acute Myocardial Infarction (ICD9: 410), Cardiac Dysrhythmias (ICD9: 427), Heart Failure (ICD9: 428), Acute 
	Pulmonary Heart Disease (ICD9: 415), Diabetes Mellitus with Complications (ICD9: 250.1-250.4, 250.6-250.9), Obesity (ICD9: 278.0)  \\ 
	[0.5em]
	Procedures CPT & 3 & Cardiovascular Procedures (including CPT 93501, 93503, 93505, etc.), Surgical Procedures on the Arteries and Vein 
	(including CPT 35686, 35501, 35509, etc.), Surgical Procedures on the Heart and Pericardium (including CPT 33533, 33534, 33535) \\ 
	[0.5em]
	Procedures ICD9 & 4 & Operations on the Cardiovascular System (ICD9: 35-39.99), Cardiac Stress Test and pacemaker checks (ICD9: 89.4), 
	Angiocardiography and Aortography (ICD9: 88.5), Diagnostic
        Ultrasound of Heart (ICD9: 88.72) \\ [0.5em] 
	Vitals & 2 & Diastolic Blood Pressure, Systolic Blood Pressure
        \\ [0.5em]
	Lab Tests & 4 & CPK (Creatine phosphokinase) (LOINC:2157-6), CRP Cardio (C-reactive protein) (LOINC:30522-7), Direct LDL (Low-density 
	lipoprotein) (LOINC:2574-2), HDL (High-Density Lipoprotein)
        (LOINC:9830-1)  \\ [0.5em] 
	Tobacco & 2 & Current Cigarette Use, Ever Cigarette Use  \\ [0.5em]
	Visits to the ER & 1 & Visits to the Emergency Room \\ [0.5em]
	Admissions & 17 & e.g., Heart Transplant or Implant of Heart
        Assist System (MSDRG: 001, 002), Cardiac Valve and Other Major 
	Cardiothoracic procedures (MSDRG: 216-221), Coronary Bypass (MSDRG: 231-234), Acute Myocardial Infarction (MSDRG: 280-285), Heart 
	Failure and Shock (MSDRG: 291-293), Cardiac Arrest (MSDRG: 296-298), Chest Pain (MSDRG: 313), Respiratory System related admissions 
	(MSDRG: 175-176, 190-192)\\  
	\bottomrule
\end{tabular}
\end{center}
\end{table}

\begin{table}[hbt]
\caption{Medical Factors in the Diabetes Dataset.}
\label{table:diabmedfactors}
\begin{center}
\begin{tabular}{p{2cm} p{8cm}} 
  \toprule
  \textbf{Ontology}   & \textbf{Examples}\\ 
  \midrule
  Demographics &  Sex, Age, Race, Zip Code  \\ [0.5em]
  Diagnoses &  e.g., Diabetes mellitus with complications,
  Thyroid disorders, Hypertensive disease, Pulmonary heart
  disease, Heart failure, Aneurysm, Skin infections,
  Abnormal glucose tolerance test, Family history of
  diabetes mellitus\\  
  [0.5em] 
  Procedures (CPT or ICD9) &  e.g., Procedure on single vessel,
  Insertion of intraocular lens prosthesis at time of cataract
  extraction, Venous catheterization, Hemodialysis, Transfusion of
  packed cells \\  [0.5em] 
  Admissions & e.g., Diabetes (with and without) complications, Heart
  failure and shock, Deep Vein Thrombophlebitis, Renal failure, Chest
  pain, Chronic obstructive pulmonary disease, Nutritional. \& misc
  metabolic disorders, Bone Diseases \& Arthropathies, Kidney \& urinary
  tract infections, Acute myocardial infarction, O.R. procedures for
  obesity, Hypertension\\ [0.5em]  
  Service by Department & Inpatient (admit), Inpatient (observe),
  Outpatient, Emergency Room\\  
  \bottomrule
	\end{tabular}
\end{center}
\end{table}

In order to organize all the available information in a uniform way for
all patients, some preprocessing of the data is needed to summarize the
information over a time interval. Details will be discussed in the next
subsection. We will refer to the summarized information of the medical
factors over a specific time interval as {\em features}.

Each feature related to diagnoses, procedures (CPT), procedures (ICD9)
and visits to the Emergency Room is an integer count of such records for
a specific patient during the specific time interval. Zero indicates
the absence of any record. Blood pressure and lab tests features are
continuous valued.  Missing values are replaced by the average of values
of patients with a record at the same time interval.  Features related
to tobacco use are indicators of current- or past-smoker in the specific
time interval. Admission features contain the total number of days of
hospitalization over the specific time interval the feature corresponds
to. Admission records are used both to form the admission features (past
admission records) and in order to calculate the prediction variable
(existence of admission records in the target year). We treat our
problem as a classification problem and each patient is assigned a {\em
  label}: $1$ if there is a heart-related (or diabetes-related)
hospitalization in the target year and $-1$ otherwise.

\subsection{Heart Data Preprocessing}

In this section we discuss several data organization and preprocessing
choices we make for the heart dataset. For each patient, a target year
is fixed (the year in which a hospitalization prediction is sought) and
all past patient records are organized as follows.

\subsubsection{Summarization of the medical factors in the history of a
  patient} After exploring multiple alternatives, an effective way to
summarize each patient's medical history is to form four time blocks for
each medical factor. Time blocks 1, 2, and 3 summarize the medical
factors over one, two, and three years before the target year, whereas
the 4th block summarizes all earlier patient records. For tobacco use,
there are only two features, indicating whether the patient is currently
smoking and whether he/she has ever smoked. After removing features with
zero standard deviation, this process results in a vector of 212
features for each patient.

\subsubsection{Selection of the target year} As a result of the nature of
the data, the two classes are highly imbalanced. When we fix the target
year for all patients to be 2010, the number of hospitalized patients is
about 2\% of the total number of patients, which does not yield enough
positive samples for effective training. Thus, and to increase the
number of hospitalized patient examples, if a patient had only one
hospitalization throughout 2007\---2010, the year of hospitalization is
set as the target year for that patient. If a patient had multiple
hospitalizations, a target year between the first and the last
hospitalization is randomly selected.

\subsubsection{Setting the target time interval to be a year} After testing
several options, a year appears to be an appropriate time interval for
prediction. Shorter prediction windows increase variability and do not
allow sufficient time for prevention. Moreover, given that
hospitalization occurs roughly uniformly within a year, we take the
prediction time interval to be a calendar year.

\subsubsection{Removing noisy samples} Patients who have no records before
the target year are impossible to predict and are thus removed.

\subsection{Identifying Diabetes-Related
  Hospitalizations} \label{sec:diabadmis}

Identifying the hospitalizations that occur mainly due to diabetes is
not a trivial task, because for financial reasons (i.e., higher
reimbursement) many diabetes-related hospitalizations are recorded in
the system as other types of admissions, e.g., heart-related. Therefore,
as a first step, we seek to separate diabetes-related admissions from
all the rest. To that end, we consider all patients with at least one
admission record between 1/1/2007 and 12/31/2012. From this set,
patients with at least one diabetes mellitus record during the same
period are assigned to the \textit{diabetic population}, while the rest
are assigned to the \textit{non-diabetic population}.

We list the union of all unique admission types for both populations
(732 unique types). The total number of admission samples for the
diabetic and non-diabetic populations are $N_1=47,352$ and
$N_2=116,934$, respectively. For each type of admission $d$, each
admission sample can be viewed as the outcome of a binary random
variable that takes the value $1$, if the hospitalization occurs because
of this type of admission, and $0$, otherwise. Thus, we can transform
the two sets of admission records for the two populations into binary
($0$/$1$) sequences. By (statistically) comparing the proportions of $d$
in the two populations, we can infer whether admission $d$ was caused
mainly by diabetes or not.

To that end, we will utilize a statistical hypothesis test comparing
sample differences of proportions.  Suppose we generate two sets of
admissions $\scrS_1$ and $\scrS_2$ of size $N_1$ and $N_2$ drawn from
the diabetic and the non-diabetic patient populations,
respectively. Consider a specific admission type $d$ and suppose that it
appears with probability $p_1$, out of all possible admission types in
$\scrS_1$. Similarly, a type $d$ admission appears with probability
$p_2$ in $\scrS_2$. Given now the two sets of admissions from diabetics
($\scrS_1$) and non-diabetics ($\scrS_2$), let $P_1$ and $P_2$ be the
corresponding sample proportions of type $d$ admissions. We want to
statistically compare $P_1$ and $P_2$ and assess whether a type $d$
admission is more prevalent in $\scrS_1$ vs.\ $\scrS_2$. Consider as
the null hypothesis the case where $p_1=p_2$, i.e., a type $d$ admission
is equally likely in the two populations. Under the null hypothesis, the
sampling distribution of differences in proportions is approximately
normally distributed, with its mean and standard deviation given by
\begin{equation*}
\mu_{P_1-P_2}=0 \quad \mbox{and} \quad
\sigma_{P_1-P_2}=\sqrt{pq\left(\cfrac{1}{N_1}+\cfrac{1}{N_2}\right)}, 
\end{equation*}
where $p=(N_1P_1+N_2P_2)/(N_1+N_2)$ is used as an estimate of the
probability of a type $d$ admission in both populations and $q=1-p$. By
using the standardized variable $z=(P_1-P_2)/(\sigma_{P_1-P_2})$ we can
assess if the results observed in the samples differ markedly from the
results expected under the null hypothesis. We do that using the single
sided $p$-value of the statistic $z$. The smaller the $p$-value is, the
higher the confidence we have in the alternative hypothesis or
equivalently in the fact that the diabetic patients have higher chance
of getting admission records of type $d$ than the non-diabetic ones
(since we consider the difference $P_1-P_2$).  We list admission types
in increasing order of $p$-value and we set a threshold of $p$-value
$\leq \alpha=0.0001$; admission types with $p$-value less than $\alpha$
are considered to be attributed to diabetes.~\footnote{Apart from
  selecting a small-value $\alpha$, we also ensure that the cumulative
  fraction of patients that are potentially labeled as belonging to the
  hospitalized class is not too small, so that the dataset is not
  highly imbalanced.} Examples of diabetes-related admissions are shown
in Table~\ref{table:diabmedfactors}.

\subsection{Diabetes Data Preprocessing}

The features are formed as combinations of different medical factors (instead
of considering the factors as separate features) that better describe what
happened to the patients during their visits to the hospital. Specifically, we
form triplets that consist of a diagnosis, a procedure (or the information
that no procedure was done), and the service department. An example of a
complex feature (a triplet) is the diagnosis of ischemic heart disease that
led to an adjunct vascular system procedure (procedure on single vessel) while
the patient was admitted to the inpatient care. Clearly, since each category
can take one of several discrete values, a huge number of combinations should
be considered. Naturally, not all possible combinations occur, which reduces
significantly the total number of potential features that describe each
patient. Also for each patient, we extract information about the diabetes type
over their history and demographics including age, gender and race.  Next, we
present several data organization and preprocessing steps we take. For each
patient, a target year is fixed and all past patient records are organized as
follows.

\subsubsection{Forming the complex features} We create a
diagnoses-procedures indicator matrix to keep track of which diagnosis
occurs with which procedure. The procedures that are not associated with
any diabetes-related diagnosis are removed. Procedures in the dataset
are listed in the most detailed level of the ICD9 coding system
\cite{icd9} or the CPT coding system \cite{cpt}. We group together
procedures that belong to the same ICD9/CPT family, resulting in 31
categories (out of 2004 in total).

\subsubsection{Summarization of the complex features in the history of a
  patient} We use the same approach as with heart diseases: we form four time
blocks for each medical factor with all corresponding records summarized over
one, two, three years before the target year, and a fourth time block
containing averages of all the earlier records. This produces a
$9,402$-dimensional vector of features characterizing each patient.

\subsubsection{Reducing the number of features} We remove all the
features that do not contain enough information for a significant amount
of the population (less than 1\% of the patients), as they could not
help us generalize. This leaves $320$ medical and $3$ demographical
features.

\subsubsection{Identifying the diabetes type} The ICD9 code for diabetes is
assigned to category $250$ (diabetes mellitus). The fifth digit of the
diagnosis code determines the type of diabetes and whether it is uncontrolled
or not stated as uncontrolled. Thus, we have four types of diabetes diagnoses:
type II, not stated as uncontrolled (fifth digit $0$), type I, not stated as
uncontrolled (fifth digit $1$), type II or unspecified type, uncontrolled
(fifth digit $2$) and type I, uncontrolled (fifth digit $3$). Based on these
four types, we count how many records of each type each patient had in the
four time blocks before the target year, thus adding $16$ new features for
each patient.

\subsubsection{Setting the target time interval to a calendar year} Again, as with
heart diseases, we seek to predict hospitalizations in the target time
interval of a year starting on the 1st of January and ending on the 31st of
December.

\subsubsection{Selection of the target year} As a result of the nature
of the data, the two classes are highly imbalanced. To increase the
number of hospitalized patient examples, if a patient had only one
hospitalization throughout 2007--2012, the year of hospitalization will
be set as the target year. If a patient had multiple hospitalizations, a
target year between the first and the last hospitalizations will be
randomly selected. 2012 is set as the target year for patients with no
hospitalization, so that there is as much available history for them as
possible. By this policy, the ratio of hospitalized patients in the
dataset is 16.97\%.

\subsubsection{Removing patients with no record} Patients who have no records before the target year are removed, since there is nothing on which a prediction can be based. The total number of patients left is 33,122.

\subsubsection{Splitting the data into a training set and a test set randomly} As
is common in supervised machine learning, the population is randomly split
into a training and a test set. Since from a statistical point of view, all
the data points (patients' features) are drawn from the same distribution, we
do not differentiate between patients whose records appear earlier in time
than others with later time stamps. A retrospective/prospective approach
appears more often in the medical literature and is more relevant in a
clinical trial setting, rather than in our algorithmic approach. What is
critical in our setting is that for each patient prediction we make
(hospitalization/non-hospitalization in a target year), we only use that
patients' information before the target year.

\section{Performance Evaluation} \label{sec:per}

Typically, the primary goal of learning algorithms is to maximize the
prediction accuracy or equivalently minimize the error rate.  However, in the
specific medical application problem we study, the ultimate goal is to alert
and assist patients and doctors in taking further actions to prevent
hospitalizations before they occur, whenever possible. Thus, our models and
results should be accessible and easily explainable to doctors and not only
machine learning experts. With that in mind, we examine our models from two
aspects: prediction accuracy and interpretability.

The prediction accuracy is captured in two metrics: the \textit{false
  alarm rate} (how many patients were predicted to be in the positive
class, i.e., hospitalized, while they truly were not) and the
\textit{detection rate} (how many patients were predicted to be
hospitalized while they truly were). In the medical literature, the
detection rate is often referred to as \textit{sensitivity} and the term
\textit{specificity} is used for one minus the false alarm rate. Two
other terms that are commonly used are the {\em recall rate}, which is
the same as the detection rate, and the {\em precision rate}, which is
defined as the ratio of true positives (hospitalizations) over all the
predicted positives (true and false).  For a binary classification
system, the evaluation of the performance is typically illustrated with
the \textit{Receiver Operating Characteristic (ROC)} curve, which plots
the detection rate versus the false alarm rate at various threshold
settings. To summarize the ROC curve and be able to compare different
methods using only one metric, we will use the {\em Area Under the ROC
  Curve (AUC)}. An ideal classifier achieves an AUC equal to $1$ (or
$100\%$), while a classifier that makes random classification decisions
achieves an AUC equal to $0.5$ (or $50\%$). Thus, the ``best'' (most
accurate) classification method will be the one that achieves the
highest AUC.

For the heart study we conduct, we will also generate the ROC curve
based on patients' 10-year risk of general cardiovascular disease
derived by the Framingham Heart Study (FHS) \cite{d2008general}. FHS is
a seminal study on heart diseases that has developed a set of risk
factors for various heart problems.  The 10-year risk we are using is
the closest to our purpose and has been widely used. It uses the
following features (predictors): age, diabetes, smoking, treated and
untreated systolic blood pressure, total cholesterol, High-Density
Lipoprotein (HDL), and BMI (Body Mass Index) which can be used to
replace lipids in a simpler model. We calculate this risk value (which
we call the \textit{Framingham Risk Factor-FRF}) for every patient and
make the classification based on this risk factor only. We also generate
an ROC curve by applying random forests just to the features involved in
FRF. The generated ROC curve serves as a baseline for comparing our
methods to classifiers that are based on features suggested only by
medical intuition.

For the diabetes study, we also consider baseline classifiers that are
based only on features commonly considered by physicians.  More
specifically, the features we select are: age, race, gender, average
over the entire patient history of the hemoglobin A1c, or HbA1c for
short (which measures average blood sugar concentrations for the
preceding two to three months), and the average number of emergency room
visits over the entire patient history. All these features are part of a
$3$-year risk of diabetes metric in \cite{mcallister2014stress}.  We
apply random forests to just these features to obtain a baseline to
compare our methods against.

Let us also note that we will compare our new algorithm ACC to SVMs (linear
and RBF), and two other hierarchical approaches that combine clustering with
classification, to which we refer as Cluster-Then-Linear-SVM (CT-LSVM) and
Cluster-Then-Sparse-Linear-SVM (CT-SLSVM). Specifically, CT-LSVM first
clusters the positive samples (still based on the feature set $\mathcal{C}$)
with the widely used $k$-means method~\cite{friedman2001elements}, then copies
negative samples into each cluster, and finally trains classifiers with linear
SVM for each cluster. The only difference between algorithm CT-SLSVM and
CT-LSVM is that CT-SLSVM adopts sparse linear SVM in the last step.

Notice that ACC implements an alternating procedure while CT-LSVM, CT-SLSVM do
not. With only one-time clustering, CT-LSVM and CT-SLSVM create unsupervised
clusters without making use of the negative samples, whereas ACC is taking
class information and classifiers under consideration so that the clusters
also help the classification.

\section{Experimental Results} \label{sec:results}

In this section, we will present experimental results on the two datasets for
all methods we have presented so far, in terms of both accuracy and
interpretability.

For SVM, tuning parameters are the misclassification penalty coefficient
$C$ (cf. (\ref{svm_sparse})) and the kernel parameter $\sigma$; we used
the values $\{0.3, 1, 3\}$ and $\{0.5, 1, 2, 7, 15, 25, 35, 50,
70, 100\}$, respectively.  Optimal values of $1$ and $7$, respectively,
were selected by cross-validation.

For $K$-LRT, we quantize the data as shown in Table~\ref{table:featquant}.
\begin{table}
	\caption{Quantization of Features.}
	\label{table:featquant}
\begin{center}
        	\begin{tabular}{p{3.2cm} p{2cm} p{7cm}} 
		\toprule
		\textbf{Features} & \textbf{Levels of quantization} & \textbf{Comments}\\ 
		\midrule
		Sex & 3 & 0 represents missing information  \\ [0.5em]
		Age & 6 & Thresholds at 40, 55, 65, 75 and 85 years old
                \\ [0.5em] 
		Race & 10 &   \\ [0.5em]
		Zip Code & 0 & Removed due to its vast variation \\ [0.5em]
		Tobacco (Current and Ever Cigarette Use) & 2 & Indicators of tobacco use  \\ [0.5em]
		Diastolic Blood Pres-sure (DBP) & 3 & Level 1 if DBP $< 60$mmHg, Level 2 if $60$mmHg $\leq$ DBP $\leq$ $90$mmHg and Level 3 if DBP $>$ $90$mmHg  \\ [0.5em]
		Systolic Blood Pressure (SBP) & 3 & Level 1 if SBP $< 90$mmHg, Level 2 if $90$mmHg $\leq$ SBP $\leq$ $140$mmHg and Level 3 if SBP $>$ $140$mmHg  \\ [0.5em]
		Lab Tests & 2 & Existing lab record or Non-Existing lab record in the specific time period  \\ [0.5em]
		All other dimensions & 7 & Thresholds are set to 0.01\%, 5\%, 10\%, 20\%, 40\% and 70\% of the maximum value of each dimension  \\ [0.5em]
		\bottomrule
	\end{tabular}
\end{center}
\end{table}
After experimentation, the best performance of $K$-LRT is achieved by setting
$k=4$.

\begin{figure}[ht]
\centering
\includegraphics[width=0.98\columnwidth]{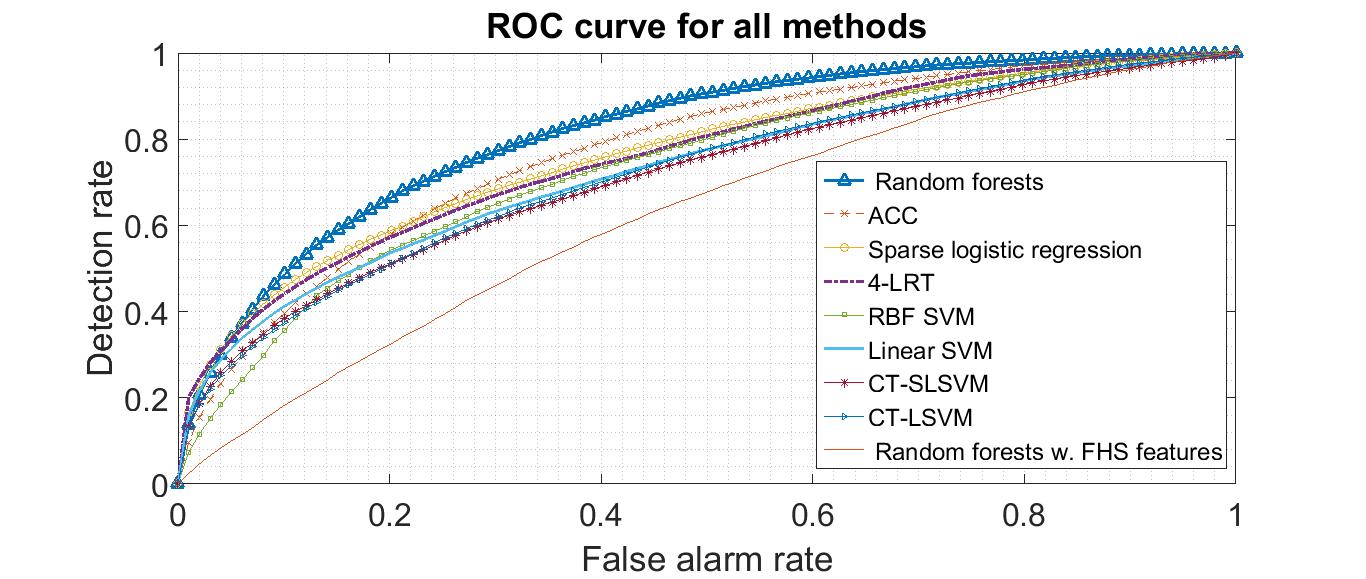}
\caption{ROC curves for the heart data.}
\label{figure:rocheart}
\end{figure}

\begin{figure}[ht]
	\centering
	\includegraphics[width=0.98\columnwidth]{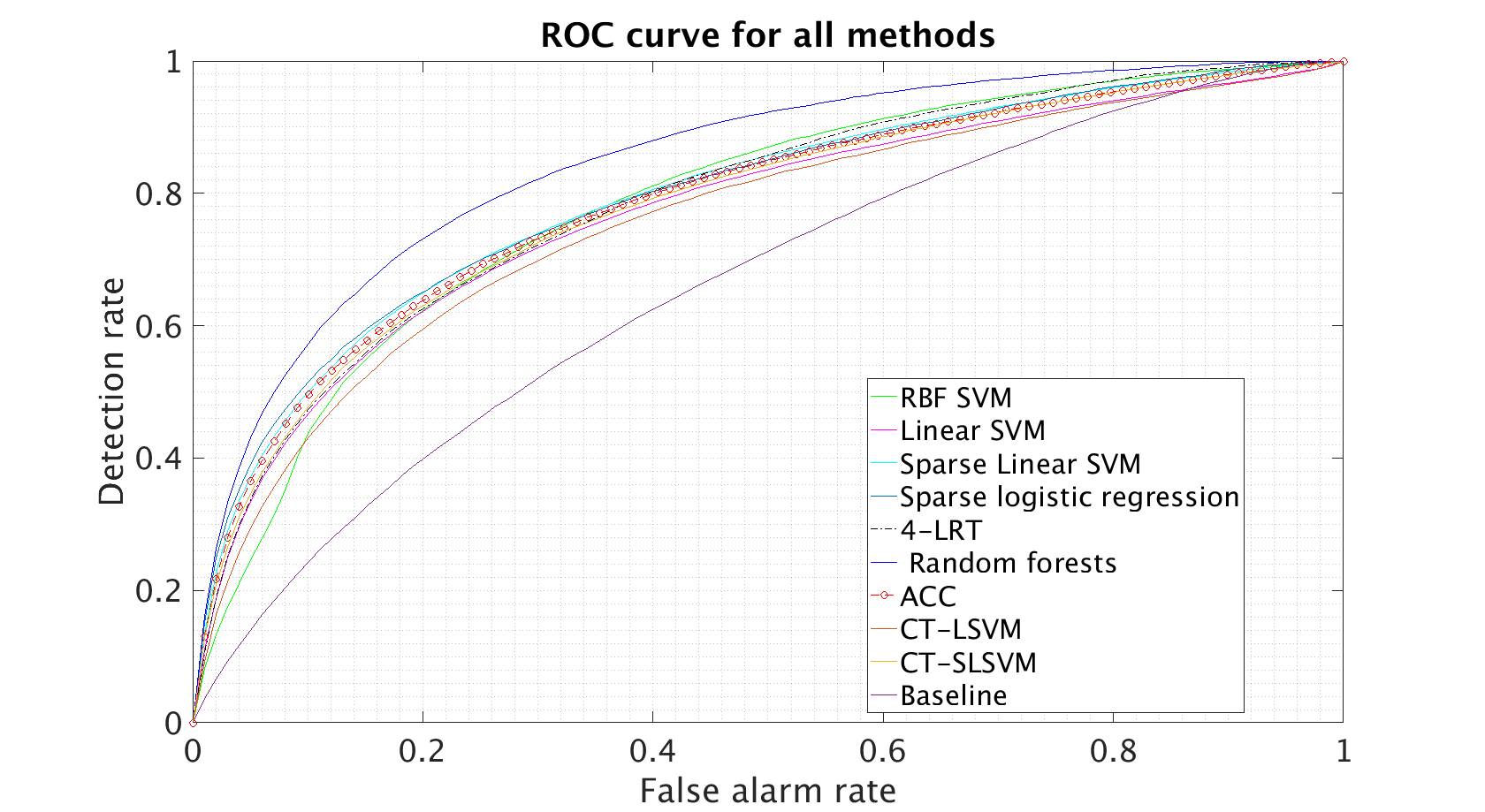} 
	\caption{ROC curves for diabetes data.}
	\label{figure:rocdiabetes}
\end{figure}

In Figures~\ref{figure:rocheart} and \ref{figure:rocdiabetes}, we
present the ROC curves of all methods, for a particular random
  split of the data into a training and test set. In
Tables~\ref{table:heartdata} and \ref{table:diabetesdata}, we present
the average (avg.) and the standard deviation (std) of the AUC over 10
different splits of the data into a training and a test set. In these
tables, Lin.\ and RBF SVM correspond to SVM with a linear and an RBF
kernel, respectively. Sparse LR corresponds to sparse logistic
regression (cf. Sec.~\ref{sec:LR}). FRF $10$-yr risk corresponds to
thresholding the Framingham $10$-year risk and random forests on FRF
features simply trains a random forest on the features used in the
Framingham $10$-year risk. We also report the baseline diabetes method
we presented in the previous subsection in the last row of
Table~\ref{table:diabetesdata}.

\begin{table}[htbp]
	\centering
	\caption{\small Prediction accuracy (AUC) on heart data.}
        	\label{table:heartdata}
			\begin{tabular}{ccc}
				\toprule
				Settings & avg. AUC & std AUC  \\  
				\midrule
				ACC, $L=1$ (SLSVM) & 76.54\% & 0.59\%  \\
				ACC, $L=2$ & 76.83\% & 0.87\%  \\ 
				ACC, $L=3$ & 77.06\% & 1.04\%  \\ 
				ACC, $L=4$ & 75.14\% & 0.92\%  \\ 
				ACC, $L=5$ & 75.14\% & 1.00\%  \\ 
				ACC, $L=6$ & 74.32\% & 0.87\%  \\ 
				$4$-LRT & 75.78\% & 0.53\%  \\ 
				Lin. SVM & 72.83\% & 0.51\%  \\ 
				RBF SVM & 73.35\% & 1.07\%  \\
				sparse LR & 75.87\% & 0.67\%  \\
				CT-LSVM ($L=2$) & 71.31\% & 0.76\%  \\
				CT-SLSVM ($L=2$) & 71.97\% & 0.73\% \\
				random forests & 81.62\% & 0.37\% \\
                                FRF $10$-yr risk & 56.48\% & 1.09\%\\
				random forests on FRF features & 62.20\%
                                & 1.13\% 
                                \\ 
				\bottomrule
			\end{tabular}  	
\end{table}

\begin{table}[htbp]
	\centering
	\caption{Prediction Accuracy (AUC) on diabetes data.}
\label{table:diabetesdata}
			\begin{tabular}{ccc}
				\toprule
				Settings & avg. AUC & std AUC  \\  
				\midrule
				ACC, $L=1$ (SLSVM) & 79.24\% & 0.52\%  \\
				ACC, $L=2$ & 78.55\% & 0.41\%  \\ 
				ACC, $L=3$ & 78.53\% & 0.41\%  \\ 
				ACC, $L=4$ & 78.46\% & 0.35\%  \\ 
				ACC, $L=5$ & 78.36\% & 0.36\%  \\ 
				ACC, $L=6$ & 78.18\% & 0.50\%  \\ 
				$4$-LRT & 78.74\% & 0.28\%  \\ 
				Lin. SVM & 76.87\% & 0.48\%  \\
				RBF SVM & 77.96\% & 0.27\%  \\
				sparse LR & 78.91\% & 0.38\%  \\
				CT-LSVM ($L=2$) & 75.63\% & 0.50\%  \\
				CT-SLSVM ($L=2$) & 77.99\% & 0.49\% \\
				random forests & 84.53\% & 0.26\% \\
				random forests on selected features
                                  (baseline)  & 65.77\% & 0.47\% \\
				\bottomrule
			\end{tabular}  	
\end{table}

Based on the results, random forests perform the best followed by our
ACC. It is interesting that using features selected by physicians (as in
FRF or the diabetes baseline method) leads to significantly inferior
performance even if a very sophisticated classifier (like random
forests) is being used. This suggests that the most intuitive medical
features do not contain all the information that could be used in making
an accurate prediction.

In terms of interpretability, with RBF SVM, the features are mapped
through a kernel function from the original space into a
higher-dimensional space. This, however, makes the features in the new
space not interpretable. Random forests are also not easy to
interpret. While a single tree classifier which is used as the base
learner is explainable, the weighted sum of a large number of trees
makes it relatively complicated to find the direct attribution of each
feature to the final decision. LRT itself lacks interpretability,
because we have more than 200 features for each sample and there is no
direct relationship between prediction of hospitalization and the
reasons that led to it. On the other hand, sparse linear SVM (SLSVM
which coincides with ACC using $L=1$ cluster), ACC, $K$-LRT, and sparse
LR are easily interpretable because they are based on sparse classifiers
involving relatively few features. ACC, in addition, clusters patients
and cluster membership provides extra interpretation. 

Our modified LRT, $K$-LRT, is particularly interpretable and it is
surprising that such a simple classifier has strong performance.  It
highlights the top $K$ features that lead to the classification
decision.  These features could be of help in assisting physicians
reviewing the patient's EHR profile and formulating
hospitalization-prevention strategies. To provide an example of
intuition that can be gleaned from this information, we consider the
heart disease dataset and in Table~\ref{table:top10feats_1LRT} we
present the features highlighted by $1$-LRT. We remind the reader that
in $1$-LRT, each test patient is essentially associated with a single
feature. For each feature $j$, we $(i)$ count how many times it was
  selected as the primary feature in the test set, and $(ii)$ calculate
  the average likelihood ratio $p(z_j|y=1)/p(z_j|y=-1)$ over all test
  patients. We normalize both quantities $(i)$ and $(ii)$ to have zero
  mean and variance equal to $1$. The average of these two normalized
  quantities is treated as the {\em importance score} of the feature
  $j$. We rank the importance scores and report the top $10$ features in
  Table~\ref{table:top10feats_1LRT}. In the table, CPK stands for
creatine phosphokinase, an enzyme which, when elevated, it indicates
injury or stress to the heart muscle tissue, e.g., as a result of a
myocardial infarction (heart attack). It is interesting that in
  addition to heart-related medical factors, utilization features such
  as lab tests and emergency room visits, contribute to the
  classification decision. This is likely the reason why our methods,
  which use the entirety of the EHR, perform much better than the
  Framingham-based methods.
\begin{table}[ht]
	\caption{Top 10 significant features for $1$-LRT.}
	\label{table:top10feats_1LRT}
\begin{center}
	\begin{tabular}{lp{7cm}} 
		\toprule
		\textbf{$1$-LRT} & \textbf{$1$-LRT} \\ 
		\textbf{Importance Score} & \textbf{Feature Name}\\ 
		\midrule
		10.50  & Admission of heart failure, 1 year before the target year \\ [0.5em]
		9.71  & Age \\ [0.5em]
		6.23  & Diagnosis of heart failure, 1 year before the target year\\ [0.5em]
		5.43  & Admission with other circulatory system diagnoses, 1 year before the target year \\[0.5em]
		4.38  & Diagnosis of heart failure, 2 years before the target year \\ [0.5em]
		4.16  & Diagnosis of hematologic disease, 1 year before the target year\\ [0.5em]
		3.45  & Diagnosis of diabetes mellitus w/o complications, 1 year before the target year \\ [0.5em]
		3.40  &Symptoms involving respiratory system and other chest symp-toms, 1 year before the target year \\ [0.5em]
		3.24  & visit to the Emergency Room, 1 year before the target year\\[0.5em]
		3.13  & Lab test CPK, 1 year before the target year \\ [0.5em]
		\bottomrule
	\end{tabular}
\end{center}
\end{table}

\begin{figure}[ht]
	\begin{center}
		\includegraphics[width=0.95\linewidth]{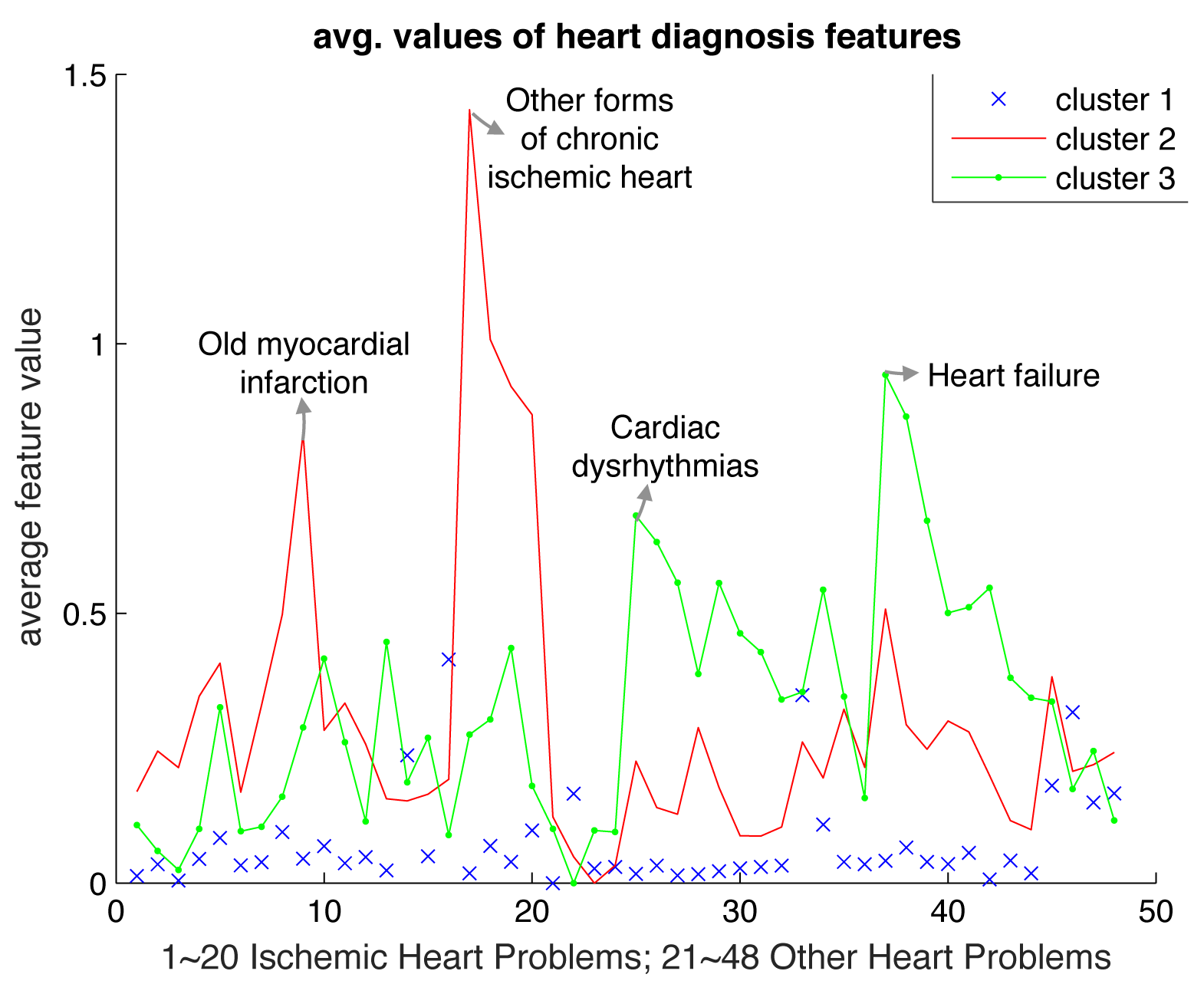} 
	\end{center}
	\caption{Average feature values in each cluster ($L=3$) for the
          heart diseases dataset.}
	\label{figure:clusterMeans}
\end{figure}

To interpret the clusters generated by ACC for the heart study (for the
case $L=3$ which yields the best performance), we plot in
Figure~\ref{figure:clusterMeans} the mean value over each cluster of
each element in the feature vector $\bx_{\mathcal{C}}$. The $3$ clusters
are well-separated. Cluster~2 contains patients with other forms of
chronic ischemic disease (mainly coronary atherosclerosis) and
myocardial infarction that had occurred sometime in the past. Cluster~3
contains patients with dysrhythmias and heart failure. Cardiologists
would agree that these clusters contain patients with very different
types of heart disease. Finally, Cluster~1 contains all other cases with
some peaks corresponding to endocardium/pericardium disease. It is
interesting, and a bit surprising, that ACC identifies meaningful
clusters of heart-disease even though it is completely agnostic of
medical knowledge.

\begin{figure}[ht]
	\begin{center}
          \includegraphics[width=0.98\linewidth]{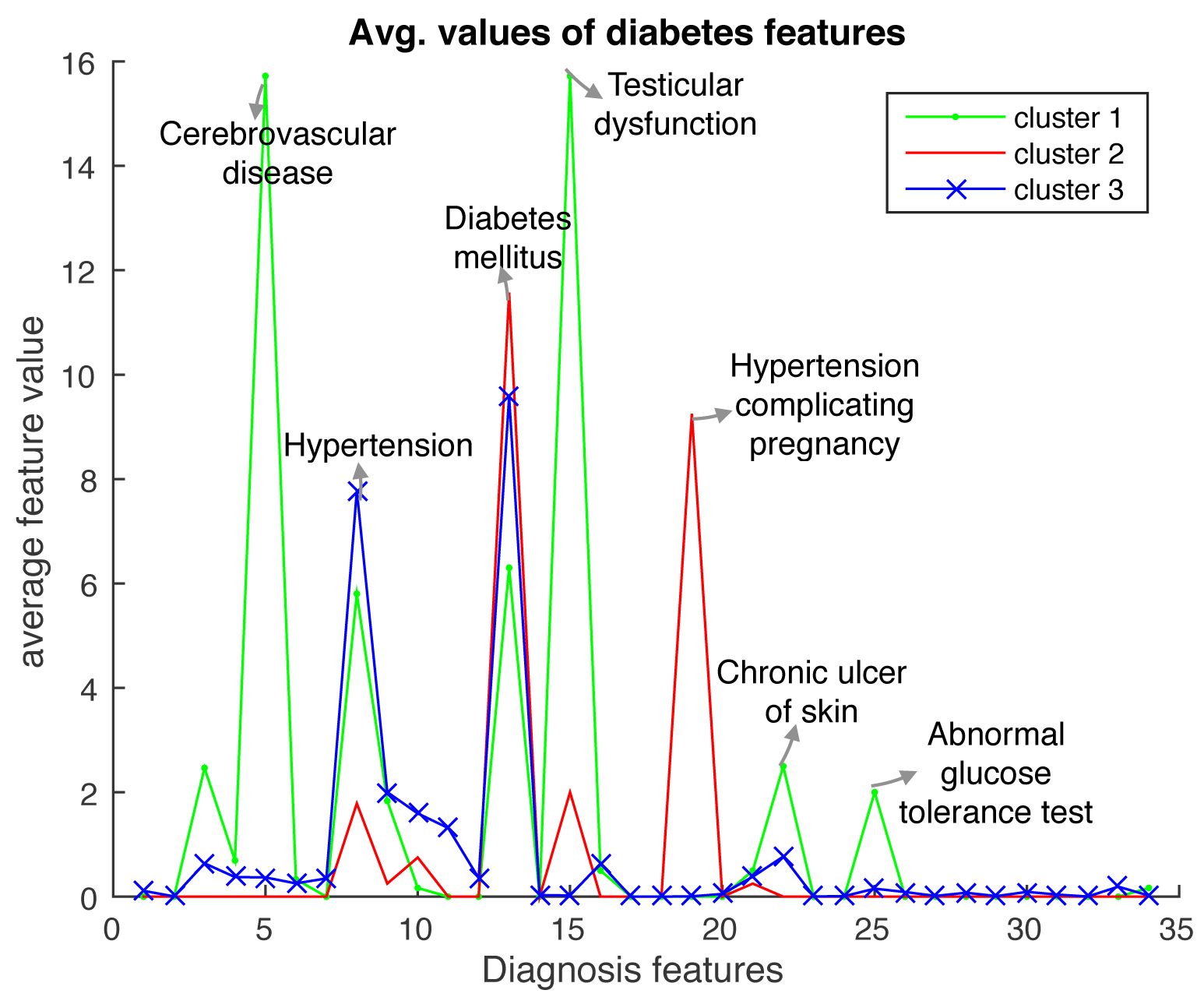}
	\end{center}
	\caption{Average feature values in each cluster ($L=3$) for the
          diabetes dataset.}
	\label{figure:diabclusterMeans}
\end{figure}

In the diabetes dataset, best ACC performance is obtained for $L=1$ (a single
cluster). Still, it is of interest to examine whether meaningful clusters
emerge for $L>1$. We plot again in Figure~\ref{figure:diabclusterMeans} the
mean value over each cluster of each element in the feature vector, using as
``diagnostic'' features the subset of features which have a correlation larger
than $0.01$ with the labels in the training set. This is done for a single
repetition of the experiment and $L=3$, yielding interesting clusters and
highlighting the interpretative power of ACC. We observe that Cluster~1
contains diabetes patients with chronic cerebrovascular disease, skin ulcers,
hypertension, an abnormal glucose tolerance test, and other complications as a
result of diabetes. Cluster~2 contains patients with diabetes complicating
pregnancy. Cluster~3 contains patients with less acute disease, combining
diabetes with hypertension. The feature values of these three clusters clearly
separate from the feature values in the negative class.

\section{Conclusions} \label{sec:concl} 

In this paper, we focused on the challenge of predicting future
hospitalizations for patients with heart problems or diabetes, based on
their Electronic Health Records (EHRs). We explored a diverse set of
methods, namely kernelized, linear and $\ell_1$-regularized linear
Support Vector Machines, $\ell_1$-regularized logistic regression and
random forests. We proposed a likelihood ratio test-based method,
$K$-LRT, that is able to identify the $K$ most significant features for
each patient that lead to hospitalization.

Our main contribution is the introduction of a novel joint clustering and
classification method that discovers hidden clusters in the positive
samples (hospitalized) and identifies sparse classifiers for each
cluster separating the positive samples from the negative ones
(non-hospitalized). The joint problem is non-convex (formulated as an
integer optimization problem); still we developed an alternating
optimization approach (termed ACC) that can solve very large instances.
We established the convergence of ACC, characterized its sample
complexity, and derived a bound on VC dimension that leads to
out-of-sample performance guarantees.  

For all the methods we proposed, we evaluated their performance in terms
of classification accuracy and interpretability, an equally crucial
criterion in the medical domain. Our ACC approach yielded the best
performance among methods that are amenable to an interpretation (or
explanation) of the prediction.  

Our findings highlight a number of important insights and opportunities
by offering a more targeted strategy for ``at-risk'' individuals.  Our
algorithms could easily be applied to care management reports or
EHR-based prompts and alerts with the goal of identifying individuals
who might benefit from additional care management and outreach.
Depending on available resources and economic considerations, a medical
facility can select a specific point on the ROC curve to operate
at. This is equivalent to selecting a tolerable maximum false positive
(alarm) rate, or, equivalently, a minimum specificity.  Because costs
associated with preventive actions (such as tests, medications, office
visits) are orders of magnitude lower than hospitalization costs, one
can tolerate significant false alarm rates and still save a large amount
of money in preventable hospitalization costs. To get a sense of this
difference, the average cost per hospital stay in the U.S. was \$9,700
in 2010~\cite{us-hospital-stays}, with some heart related
hospitalizations costing much more on average (e.g., \$18,200 for Acute
Myocardial Infarction). In contrast, an office visit costs on the order
of \$200, tests like an ECG or an echo on the order of \$100-\$230, and
a 90-day supply of common medication (hypertension or cholesterol) no
more than \$50. Clearly, even a small fraction of prevented
hospitalizations can lead to large savings. Our methods can be seen as
enabling such prevention efforts.

\appendix

\section{Proof of Proposition \ref{prop:mip}}  
\label{sec:appA}

\pf
Let $C^*_{JCC}$ and $C^*_{MIP}$ be the optimal objective values of
problems (\ref{opt:JCC}) and (\ref{eqn:MIP}).
	
Given any feasible solution to the JCC problem (\ref{opt:JCC}): $l(i),
\boldsymbol{\beta}^l, \beta_0^l, \zeta_i^l, \forall l, i$, and
$\xi_{i,JCC}^l(i)$, a feasible solution to the MIP problem is:
\[
z_{il}=\begin{cases}
  1, ~ l(i)=l,\\
  0, ~ \text{otherwise},
	\end{cases}
        \quad \xi_{i,MIP}^l=\begin{cases}
          \xi_{i,JCC}^l, ~ l(i)=l,\\
          0, ~ \text{otherwise};
\end{cases}
\]
and $\boldsymbol{\beta}^l, \beta_0^l, \zeta_i^l$ remain the same as in
the JCC solution.
	
The feasibility of the constructed MIP solution is verified as follows.
Notice that except for the $2nd$ constraint in the MIP
formulation~(\ref{eqn:MIP}) (the big-$M$ constraint), all other
constraints can be easily verified to be satisfied by the constructed
MIP solution. For the big-$M$ constraint, if $z_{il} =1$, then $M\sum_{k
  \neq l} z_{ik}=0,$ and the big-$M$ constraint holds since
$\xi_{i,MIP}^l=\xi_{i,JCC}^l$. If, however, $z_{il} =0$, then
$M\sum_{k \neq l} z_{ik}=M,$ and the big-$M$ constraint
also holds (trivially).
	
The above two feasible solutions have the same objective values, and
this equality holds for any feasible solution to the JCC problem, hence
we can conclude that $C^*_{JCC}\geq C^*_{MIP}$.
	
Next, we prove that each optimal solution to the MIP problem satisfies
$\xi_{i,MIP}^l=0$ when $z_{il}=0.$ Note that when $z_{il}=0$,
$M\sum_{k \neq l} z_{ik}=M,$ and the big-$M$ constraint
becomes $\xi_{i,MIP}^l \geq
1-y_i^+\beta_0^l-\sum_{d=1}^{D}y_i^+\beta_d^l x^+_{i,d}-M $, which will
always hold since $M$ is a large enough number. Therefore, to minimize
the objective, the optimal solution should select the smallest feasible
$\xi_{i,MIP}^l$, i.e., $\xi_{i,MIP}^l=0$.
	
Given an optimal solution to the MIP problem, a corresponding feasible
solution to JCC problem is: if $z_{il}=1$, then $\xi_{i,JCC}^l =
\xi_{i,MIP}^l, \text{~and~}l(i)=l;$ and all other variables retain their
values in the MIP solution. Since the two solutions have the same
objective cost, it follows $C^*_{JCC}\leq C^*_{MIP}.$
\qed

\section{Proof of Theorem \ref{thm:samplecomplexity}} \label{sec:appB}

\pf
To simplify notation we drop the cluster index $l$. We will use a result
from~\cite{bousquet2004introduction}. We note that the family of linear
classifiers in a $D$-dimensional space has VC-dimension $D+1$
(\cite{vapnik}). Let $\mathcal{G}$ be a function family
with VC-dimension $D+1$. Let $R_{N}(g)$ denote the training error rate
of classifier $g$ on $N$ training samples randomly drawn from an
underlying distribution $\mathcal{P}$. Let $R(g)$ denote the expected
test error of $g$ with respect to $\mathcal{P}$. The following theorem
from \cite{bousquet2004introduction} is useful in establishing our
result.
\begin{thm}[\cite{bousquet2004introduction}]\label{thm:bousquet}
  If the function family $\mathcal{G}$ has VC-dimension $D+1$, then the
  probability
  \begin{multline} \label{bousq} P\left[ R(g)-R_{N}(g) \leq 
      2\sqrt{2\frac{(D+1)\log\frac{2eN}{D+1}+ \log\frac{2}{\rho}}{N}}\
    \right] \\ \geq 1-\rho
\end{multline}
for any function $g\in \mathcal{G}$ and $\rho\in(0,1)$.
\end{thm}

For the given $\epsilon$ in the statement of
Theorem~\ref{thm:samplecomplexity}, select large enough $N$ such that
\[
\epsilon \geq
2\sqrt{2\frac{(D+1)\log\frac{2eN}{D+1}+\log\frac{2}{\rho}}{N}}, 
\]
or 
\begin{equation} \label{sa1}
\frac{2}{\rho} \leq \exp\left\{ \frac{N\epsilon^2}{8} - (D+1)
  \log\left(\frac{2eN}{D+1}\right)\right\}.  
\end{equation} 
It follows from Thm.~\ref{thm:bousquet}, 
\begin{equation} \label{sa2}
P\left[ R(g)-R_{N}(g) \geq \epsilon\right] \leq \rho. 
\end{equation} 
In our setting, the classifier $g$ is restricted to a $Q$-dimensional
subspace of the $D$-dimensional feature space. Thus, the bound in
(\ref{sa1}) holds by replacing $D$ with $Q$ in the right hand side and
the bound in (\ref{sa2}) holds for any such $Q$-dimensional subspace
selected by the $\ell_1$-penalized optimization. Since there are
$D\choose Q$ possible choices for the subspace, using the union bound we
obtain:
\[
P\left[ R(g)-R_{N}(g) \geq \epsilon\right] \leq 
{D\choose Q} \rho. 
\]
Using the bound ${D\choose Q} \leq (\frac{eD}{Q})^{Q} =
\exp(Q\log\frac{eD}{Q})$, it follows:  
\begin{equation} \label{sa3} 
P\left[ R(g)-R_{N}(g) \geq \epsilon\right] \leq
\rho \exp\left\{Q\log\frac{eD}{Q}\right\}. 
\end{equation}
For the given $\delta\in (0,1)$ in the statement of
Theorem~\ref{thm:samplecomplexity}, select small enough $\rho$ such 
that
\[
 \delta \geq \rho \exp\left\{Q\log\frac{eD}{Q}\right\},  
\]
or equivalently
\[
\frac{1}{\delta} \leq \frac{1}{\rho}
\exp\left\{-Q\log\frac{eD}{Q}\right\}.   
\]
Using (\ref{sa1}) (with $Q$ replacing $D$), we obtain
\[ 
\log\frac{2}{\delta} \leq \frac{N\epsilon^{2}}{8} - (Q+1)
  \log\left(\frac{2eN}{Q+1}\right) -Q\log\frac{eD}{Q},  
\] 
which implies that $N$ must be large enough to satisfy
\[ 
N \geq \frac{8}{\epsilon^{2}} \left[ \log\frac{2}{\delta} 
+ (Q+1)\log\frac{2eN}{Q+1} + Q\log\frac{eD}{Q} \right]. 
\]
This establishes $P\left( R(g)-R_{N}(g) \geq \epsilon\right) \leq
\delta$, which is equivalent to Theorem~\ref{thm:samplecomplexity}.
\qed

\section{Proof of Theorem \ref{thm:convergence}}
\label{sec:appC}

\pf
  At each alternating cycle, and for each cluster $l$, we train a SLSVM
  using as training samples the positive samples of that cluster
  combined with all negative samples. This produces an optimal value
  $O^{l}$ for the corresponding SLSVM training optimization problem
  (cf.~(\ref{opt:JCC})) and the corresponding classifier
  $(\boldsymbol{\beta}^{l}, \beta^{l}_{0})$. Specifically, the SLSVM
  training problem for cluster $l$ is:
\begin{equation}\label{opt:SLSVM}
\begin{array}{rl}
  O^{l}=\min_{\substack{\bbeta^{l},\beta^{l}_0,\\
      \zeta_{j}^{l},\xi_{i}^{l}}} & \frac{1}{2}||\bbeta^{l}||^2 +
  \lambda^{+}\sum_{i=1}^{N^{+}_{l}} \xi_{i}^{l} +
  \lambda^{-}\sum\limits_{j=1}^{N^{-}} \zeta_{j}^{l}\\
\text{s.t.}& \xi_i^{l} \geq 1-y^{+}_i \beta^{l}_0 - \sum_{d=1}^D  y^{+}_i \beta^{l}_{d} x^{+}_{i,d},\ \forall i,\\
& \zeta_j^{l} \geq 1-y^{-}_j \beta^{l}_0 - \sum_{d=1}^D  y^{-}_j
		\beta^{l}_{d} x^{-}_{j,d},\  \forall j,\\

& \sum_{d=1}^D |\beta^{l}_{d}| \leq T^l,\ \xi_i^{l}, \zeta_{j}^{l}\geq
0,\ \forall i, j. 
\end{array}
\end{equation}
Set
\begin{equation*}
  Z  = \sum_{l=1}^{L} O^{l} = \sum_{l=1}^{L}
  \bigg(\frac{1}{2}||\boldsymbol{\beta}^{l}||^2 +
  \lambda^{-}\sum_{j=1}^{N^{-}} \zeta^{l}_{j}\bigg) +
  \lambda^{+}\sum_{i=1}^{N^{+}} \xi^{l(i)}_{i},
 \end{equation*}
 where $l(i)$ maps sample $i$ to cluster $l(i)$, $\sum_{l=1}^{L}
 N_{l}^{+}= N^{+}$, and $\boldsymbol{\beta}^{l}$, $\beta^{l}_{0}$,
 $\zeta^{l}_{j}$, and $\xi^{l(i)}_{i}$ are optimal solutions of
 (\ref{opt:SLSVM}) for each $l$. Let us now consider the change of $Z$
 at each iteration of the ACC training procedure.
	
 First, we consider the re-clustering step
 (Alg.~\ref{alg:reclustering2}) given computed SLSVMs for each
 cluster. During the re-clustering step, the classifier and slack
 variables for negative samples are not modified. Only the
 $\xi^{l(i)}_{i}$ get modified since the assignment functions $l(i)$
 change. When we switch positive sample $i$ from cluster $l(i)$ to
 $l^{*}(i)$, we can simply assign value $\xi^{l(i)}_{i}$ to
 $\xi^{l^*(i)}_{i}$.  Therefore, the value of $Z$ does not change during
 the re-clustering phase and takes the form
\begin{equation*}
Z= \sum_{l=1}^{L} \bigg(\frac{1}{2}||\boldsymbol{\beta}^{l}||^2 
+ \lambda^{+}\sum_{\{i: l^{*}(i)=l\}} \xi^{l}_{i}
+ \lambda^{-}\sum_{j=1}^{N^{-}} \zeta^{l}_{j}\bigg).
\end{equation*}
	
Next, given new cluster assignments, we re-train the local classifiers by
resolving problem (\ref{opt:SLSVM}) for each cluster $l$. Notice that
re-clustering was done subject to the constraint in
Eq.~(\ref{eqn:impconstr}). Since $y^{+}_{i}=1$ for all positive samples,
we have
\begin{align*}
\xi^{l(i)}_i \geq & 1- \beta^{l(i)}_0 - \sum\limits_{d=1}^D
	\beta^{l(i)}_{d} x^{+}_{i,d} \\
	\geq & 1- \beta^{l^{*}(i)}_0 - \sum\limits_{d=1}^D  
	\beta^{l^{*}(i)}_{d} x^{+}_{i,d}. 
\end{align*}
The first inequality is due to $\xi^{l(i)}_i$ being feasible for
(\ref{opt:SLSVM}). The second inequality is due to $y^{+}_{i}=1$ and
Eq.~(\ref{eqn:impconstr}). Thus, by assigning $\xi^{l(i)}_{i}$ to
$\xi^{l^*(i)}_{i}$ it follows that the $\xi^{l^*(i)}_{i}$ remain
feasible for problem (\ref{opt:SLSVM}). Given that the remaining
decision variables do not change, $(\boldsymbol{\beta}^{l},
\beta^{l}_{0}, \zeta^{l}_{j}, \xi^{l^*(i)}_{i},\ \forall
i=1,\ldots,N^+_l,\ \forall j=1,\ldots,N^-)$ forms a feasible solution of
problem (\ref{opt:SLSVM}). This solution has a cost equal to
$O^l$. Re-optimizing can produce an optimal value that is no worse. It
follows that in every iteration of ACC, $Z$ is monotonically
non-increasing. Monotonicity and the fact that $Z$ is bounded below by
zero, suffices to establish convergence.
\qed

\section{Proof of Theorem \ref{thm:VCdim}} 
\label{sec:appD} 

\pf
The proof is based on Lemma~2 of~\cite{sontag:98}. Given an assignment of
each positive sample $i$ to cluster $l(i)$, define $L$ clustering functions 
\[ g_l(i) = \begin{cases} 1, & \text{if $l(i)=l$,}\\
               0, & \text{otherwise.}
\end{cases} 
\]
Hence, positive sample $i$ is assigned to cluster $\arg\max_l
g_l(i)$. This can be viewed as the output of $(L-1)L/2$ comparisons
between pairs of $g_{l_1}$ and $g_{l_2}$, where $1 \leq l_1 < l_2 \leq
L$.  This pairwise comparison could be further transformed into a
boolean function (i.e., $\text{sgn}(g_{l_1}-g_{l_2})$). Together with
the $L$ classifiers (one for each cluster), we have a total of
$(L+1)L/2$ boolean functions.  Among all these boolean functions, the
maximum VC-dimension is $D+1$, because at most $D$ features are being
used as input.  Therefore, by Lemma~2 of \cite{sontag:98}, the
VC-dimension of the function family $\mathcal{H}$ is bounded by
$2(\frac{(L+1)L}{2}) (D+1) \log (e \frac{(L+1)L}{2})$, or equivalently
$(L+1)L (D+1) \log (e \frac{(L+1)L}{2})$.
\qed



\bibliographystyle{IEEEtran}




\end{document}